%% file: main.tex
\documentclass[10pt,twocolumn,letterpaper]{article}

\usepackage{cvpr}              %

\input{preamble}

\definecolor{cvprblue}{rgb}{0.21,0.49,0.74}
\usepackage[pagebackref,breaklinks,colorlinks,citecolor=cvprblue]{hyperref}

\def\paperID{16816} %
\def\confName{CVPR}
\def\confYear{2024}

\title{\OURS: Deferred Diffusion for High-fidelity 3D Head Avatars}

\author{Tobias Kirschstein \qquad Simon Giebenhain \qquad Matthias Nießner\\[0.2cm]
Technical University of Munich
}

\begin{document}
\twocolumn[{
\renewcommand\twocolumn[1][]{#1}%
\maketitle
\vspace{-0.5cm}
\includegraphics[width=\textwidth]{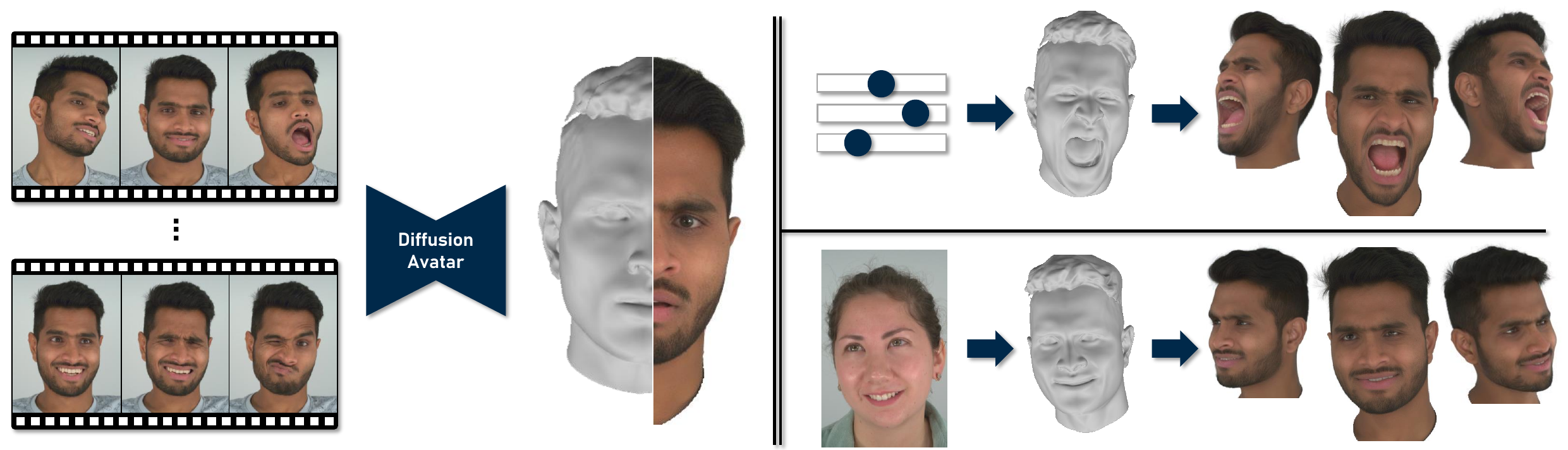}
\begin{tabularx}{\linewidth}{YY}
(a) Avatar creation & (b) Avatar animation
\end{tabularx}
\captionof{figure}{Given a set of multi-view videos and corresponding fitted meshes, we build a \textit{DiffusionAvatar} of a person. Our method translates expressions of a morphable model into realistic facial appearances of a person while also providing control over the viewpoint.
Project website: \url{https://tobias-kirschstein.github.io/diffusion-avatars/}} \vspace{2em}
\label{fig:teaser}
}]

\input{sections/0_abstract}

\input{sections/1_introduction}
\input{sections/2_related_work}
\input{sections/3_method}
\input{sections/4_results}
\input{sections/5_acknowledgements}
{
     \small
    \bibliographystyle{ieeenat_fullname}
    \bibliography{main}
}

\input{sections/X_suppl}

\end{document}

%% file: preamble.tex
\usepackage[accsupp]{axessibility} %

\usepackage[dvipsnames]{xcolor}

\usepackage{tabularx} %
\usepackage{xspace} %
\usepackage{xcolor} %
\usepackage{makecell} %
\usepackage{multirow} %
\usepackage{bm} %
\usepackage{mathtools}  %
\usepackage{colortbl} %
\usepackage{subcaption}  %

\newcommand{\OURS}{Diffusion\-Avatars}

\DeclareMathOperator*{\argmin}{arg\,min}  %

\newcolumntype{Y}{>{\centering\arraybackslash}X}
\newcolumntype{P}[1]{>{\centering\arraybackslash}p{#1}}
\newcolumntype{C}{>{\centering}X}

\newcommand{\nparticipants}{49}
\newcommand{\nresponses}{735}

%% file: sections/0_abstract.tex
\begin{abstract}

\vspace{-1em}

{\OURS} synthesizes a high-fidelity 3D head avatar of a person, offering intuitive control over both pose and expression. %
We propose a diffusion-based neural renderer that leverages generic 2D priors to produce compelling images of faces.
For coarse guidance of the expression and head pose, we render a neural parametric head model (NPHM) from the target viewpoint, which acts as a proxy geometry of the person.
Additionally, to enhance the modeling of intricate facial expressions, we condition {\OURS} directly on the expression codes obtained from NPHM via cross-attention.
Finally, to synthesize consistent surface details across different viewpoints and expressions, we rig learnable spatial features to the head's surface via TriPlane lookup in NPHM's canonical space. \\
We train {\OURS} on RGB videos and corresponding fitted NPHM meshes of a person and test the obtained avatars in both self-reenactment and animation scenarios. Our experiments demonstrate that {\OURS} generates temporally consistent and visually appealing videos for novel poses and expressions of a person, outperforming existing approaches.

\end{abstract}

%% file: sections/1_introduction.tex
\section{Introduction}

A significant problem in computer vision and graphics is the creation of photorealistic animatable avatars.
Ideally, these avatars permit consistent video renderings while providing free control over desired expression, pose, and viewpoint. 
A digital 3D head avatar may then be used in various scenarios such as VR/AR applications, immersive teleconferencing, gaming, movie animation, or as a virtual assistant.

Unfortunately, creating a 3D avatar from which one can render photorealistic images from arbitrary viewpoints is challenging.
The task becomes even more difficult when the avatar has to simultaneously offer precise control over poses and expressions, for instance given by a driving sequence from a different video or controlled manually.
Technically, this leads to a 4D photometric reconstruction problem which is typically underconstrained.
This presents animation systems with a challenging task that traditionally involves extensive efforts by 3D artists.

Many existing methods have tackled the challenge of facial image synthesis via 2D neural networks~\cite{karras2019stylegan, tov2021styleganmanipulation, hong2022dagan, richardson2021pixelstylepixel, tewari2020pie, nitzan2022mystyle}. In particular, Diffusion Models have recently been demonstrated to produce visually appealing imagery~\cite{ho2020ddpm, rombach2022ldm, salimans2022progressivedistillation, zhang2023controlnet, zeng2023attributeguideddiffusion}.
2D models are great at generating photo-realistic images but generally do not offer the degree of control and temporal consistency needed for a high-quality 3D head avatar. On the other side, methods that reconstruct an animatable 3D representation of the head~\cite{zielonka2023insta, grassal2022nha, zheng2023pointavatar, zheng2022imavatar} provide much better consistency but in many cases the resulting renderings do not have the same photo-realism as 2D models.

To this end, we introduce {\OURS}, a novel approach for 3D head avatar creation, by combining the strong image synthesis capabilities of 2D diffusion models with the view consistency of a detailed 3D head representation.
As a controllable 3D representation, we leverage the recent neural parametric head model (NPHM)~\cite{giebenhain2023nphm} due to its detailed prediction of human head geometry, which can give us better geometric cues than established mesh-based 3DMMs~\cite{FLAME:SiggraphAsia2017, paysan2009bfm}. Inspired by neural textures~\cite{thies2019dnr}, we further base face synthesis on person-specific learnable features rigged to the model's surface via NPHM's canonical space. Such surface-mapped features can effectively compensate for imperfect geometry and improve consistency across views by giving the neural renderer cues about corresponding surface points.
To avoid learning a neural renderer from scratch, we base our architecture on a pre-trained latent diffusion model (LDM)~\cite{rombach2022ldm} and convert it into an image-to-image translation model by following the ControlNet~\cite{zhang2023controlnet} paradigm to condition on the rasterized images of NPHM meshes. That way, we not only inherit a powerful image synthesis backbone but also benefit from the learned facial prior obtained from large 2D image datasets, which facilitates generalization to unseen expressions.  
Finally, we insert cross-attention blocks into the pre-trained LDM to condition the diffusion-based neural renderer on the expression codes of NPHM directly. This allows our model to produce more expressive renderings. Intuitively, the direct conditioning helps the diffusion model to distinguish subtle expression details while the rasterized NPHM renderings encode the viewpoint and the overall shape of the head. 
In summary, our contributions are as follows: 
\begin{itemize}
    \item We present {\OURS}, a diffusion-based neural renderer that leverages ControlNet to create animatable 3D head avatars.
    \item We design a method for rigging learnable spatial features to the surface of the underlying NPHM via TriPlanes.
    \item We propose direct expression conditioning via cross-attention to transfer detailed expressions from NPHM to the synthesized 3D head avatar.
\end{itemize}

%% file: sections/2_related_work.tex
\section{Related Work}

\subsection{3D Face Animation}
It is natural to approach the task of 3D Face Animation with an actual 3D representation. In the seminal work by~\cite{blanz2023morphable}, a 3D morphable model (3DMM) was introduced, which paved the way towards reconstructing animatable 3D head geometry from images or depth observations. More recent 3DMMs~\cite{FLAME:SiggraphAsia2017, paysan2009bfm, wang2022faceverse, yang2020facescape, Chai2023HiFace} still build upon this paradigm and have since been the drivers of face animation oriented applications with most methods leveraging a 3DMM to gain control over the expression and pose of the avatar. For example, Neural Head Avatars~\cite{grassal2022nha} and ROME~\cite{khakhulin2022rome} finetune the mesh topology of FLAME~\cite{FLAME:SiggraphAsia2017} to obtain more faithful and realistic mesh-based avatars. NeRFace~\cite{gafni2021nerface}, INSTA~\cite{zielonka2023insta} and RigNeRF~\cite{athar2022rignerf} build a radiance field that is controlled by a 3DMM. Other 3D representations such as points~\cite{zheng2023pointavatar} or signed distance functions~\cite{zheng2022imavatar, lin2023ssif} have also been explored for avatar creation.\\
In our work, we do not aim to learn a complete 3D representation. Instead, we utilize the 3D prior of the recent neural parametric head model (NPHM)~\cite{giebenhain2023nphm} with a 2D rendering network for high-quality image synthesis. 

\subsection{Face Synthesis in 2D}
Ever since the impactful Pix2Pix~\cite{zhu2017pix2pix} work, it has been a popular approach to obtain control over synthesized images via 2D networks. In the face synthesis domain, methods based on Generative Adversarial Networks (GANs)~\cite{goodfellow2020gan} have received a lot of attention \cite{karras2019stylegan, tov2021styleganmanipulation, hong2022dagan, richardson2021pixelstylepixel, tewari2020pie, nitzan2022mystyle}. To obtain better control over head pose and camera viewpoint, a common technique is to combine powerful 2D image synthesis networks with a controllable 3D head proxy~\cite{chan2022eg3d, sun2023next3d, tang20233dfaceshop, bergman2022gnarf, mensah2023hybridgeneratorarchitecture, deng2020discofacegan, yue2022anifacegan, thies2019dnr, lin2022face3dganinversion, tewari2020stylerig}.
Another line of work aims at directly generating consistent videos by video-to-video translation~\cite{kim2018dvp, wang2021facevid2vid} or even video stylization~\cite{yang2022Vtoonify, jamrivska2019stylizingvideoexample, fivser2017examplefaceanimation}.
Most related to our method is Deferred Neural Rendering~\cite{thies2019dnr} which allows facial animation by learning a neural renderer in order to decode neural textures rigged to Basel Face Model~\cite{paysan2009bfm}.
While 2D-based approaches produce visually appealing frames, they often struggle with synthesizing consistent images across both views and time or provide only limited viewpoint control.\\
Our method aims to improve consistency by utilizing a more powerful implicit 3D representation as geometric proxy and provides better expression generalization by utilizing a diffusion model as a synthesis backbone.

\subsection{Controllable 2D Diffusion}

Recently, Diffusion models emerged as powerful 2D image generation models. While they demonstrate superior capabilities in generating high-fidelity and diverse 2D content from text prompts~\cite{ho2020ddpm, rombach2022ldm, salimans2022progressivedistillation}, it has been a key challenge to leverage their power for other tasks.
Several works extend 2D diffusion models to text-guided video generation by employing a multi-stage approach~\cite{blattmann2023align, ho2022imagenvideo, wang2023lavie, singer2022makeavideo}, generating frames autoregressively~\cite{ho2022vdm} or stylizing an input video~\cite{wu2023tuneavideo, geyer2023tokenflow, yang2023rerender, yan2023magicprop}. 
In the human domain, diffusion models have already been applied to face editing~\cite{huang2023collaborativediffusionfaces}, animation~\cite{zeng2023attributeguideddiffusion, stypulkowski2022diffusedheads} and bodies~\cite{svitov2023dinar}.
Other works, such as ControlNet~\cite{zhang2023controlnet}, IPAdapter~\cite{ye2023ipadapter} or T2I-Adapter~\cite{mou2023t2i}, allow fine-tuning a pre-trained diffusion model on additional controls such as landmarks or depth cues, opening a wide space of possible applications for 2D face editing.
Most similar to our method, DiffusionRig~\cite{ding2023diffusionrig} proposes a diffusion model conditioned on rasterized FLAME meshes to provide 3D animation control over an avatar created from a personal photo album. 
While the synthesized images or videos of these approaches already show great controllability and visual quality, they typically lack consistency due to the absence of an underlying accurate 3D representation. \\

\subsection{Diffusion for 3D Faces}
Following a different approach, several works explore diffusion for generic 3D generation~\cite{poole2022dreamfusion, liu2023zero1to3, muller2023diffrf}. This has also been extended to 3D head generation~\cite{wang2023rodin, gu2023controllable3DDiffusion, zhang2023styleavatar3d}, text-guided 3D head editing~\cite{pan2023avatarstudio} or 3D body generation~\cite{zeng2023avatarbooth}.
Some works also generate animatable 3D asset animatable for heads~\cite{bergman2023articulated3DHead} and bodies~\cite{cao2023dreamavatar, qin2023dancingavatar, jiang2023avatarcraft} by rigging the 3D asset to a parameterized FLAME~\cite{FLAME:SiggraphAsia2017} or SMPL~\cite{loper2023smpl} mesh.
Such generated 3D avatars are view consistent by design but lack the visual quality of 2D-based diffusion models. Furthermore, animation control is limited by the underlying mesh template. In our work, we employ a more accurate, implicit 3D morphable head model for better expression control and follow ControlNet~\cite{zhang2023controlnet} and IPAdapter~\cite{ye2023ipadapter} to build a diffusion-based neural renderer for high-quality image synthesis.

%% file: sections/3_method.tex
\begin{figure*}[htb]
    \centering
    \includegraphics[width=\linewidth]{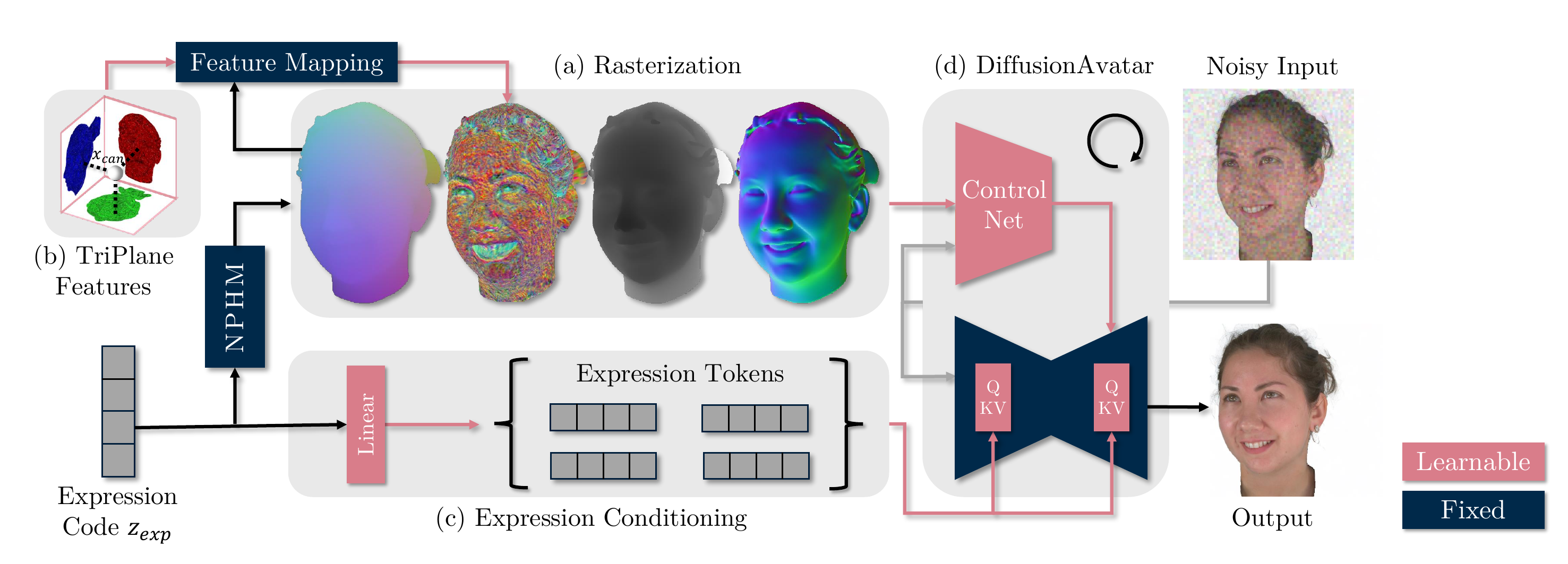}
    \vspace{-0.5cm}
    \caption{\textbf{Method overview:} We decode an NPHM expression code $z_{exp}$ in two ways to obtain a realistic image: We first extract an NPHM mesh and rasterize it from the desired viewpoint in~(a), giving us canonical coordinates, depths, and normal renderings for the head mesh. In~(b), the canonical coordinates $x_{can}$ are used to look up spatial features in a TriPlanes structure, rigging the features onto the mesh surface. Together with the rasterizer output, these mapped features form the input for the ControlNet part of DiffusionAvatar.
    The second route for the expression code goes through a linear layer depicted in~(c). It yields expression tokens that are subsequently used in a newly added cross-attention layer inside the pre-trained latent diffusion model. Intuitively, the rasterized inputs should encode pose, shape and rough expression while the direct expression conditioning hints at more detailed facial expressions. The final image is synthesized in~(d) by iteratively denoising Gaussian noise using the original DDPM denoising schedule~\cite{ho2020ddpm}.}
    \label{fig:method_overview}
\end{figure*}

\section{Method}

Our goal is to create a temporally consistent 3D head avatar of a person with explicit control over viewpoint and expression.
We approach this task by designing a diffusion-based neural renderer that decodes learnable features rigged to the surface of an implicitly defined proxy geometry.
In analogy to Deferred Neural Rendering~\cite{thies2019dnr}, we dub this approach \textit{Deferred Diffusion}.
We begin by introducing the foundations of diffusion models~(\cref{sec:Diffusion}) and NPHM~(\cref{sec:NPHM}) that we use as proxy geometry before going into detail about our method:
In the rasterization~(\cref{sec:Rasterization}) and surface feature mapping~(\cref{sec:Feature_Mapping}) stage, we generate the input images for our diffusion-based neural renderer.
These renderings encode the desired viewpoint, head shape, and global expression. For more fine-grained expression control, we add NPHM's expression codes as an additional input to the diffusion model~(\cref{sec:Expression_Conditioning}). Finally, our pipeline leverages a pre-trained diffusion model's image synthesis and generalization capabilities for better quality.

\subsection{Diffusion}
\label{sec:Diffusion}

We build our work upon latent diffusion models (LDM)~\cite{rombach2022ldm}, which operate in the the latent space of an autoencoder. Formally, LDMs employ an encoder $\mathcal{E}$ to map RGB images into a lower-dimensional latent space, facilitating generative tasks~\cite{esser2021taming}. In the following, any mention of an image $x_0$ always refers to an RGB image $I$ that was mapped into the autoencoder's latent space via $x_0 = \mathcal{E}(I)$ to obtain a latent image.

In diffusion, the fixed forward process iteratively adds noise to the latent image:
\begin{align}
    q(x_\tau|x_{\tau-1}) &= \mathcal{N}\left(x_\tau; \sqrt{\alpha_\tau}x_{\tau-1}, (1-\alpha_\tau)\mathbf{I}\right)
\end{align}
where $\tau = 1 ... N$ indicates the denoising step and the forward process variances $\alpha_\tau$ define the noise scheduling~\cite{ho2020ddpm}.
In practice, noisy samples $x_\tau$ are obtained with the standard Gaussian reparameterization:
\begin{equation}
    x_\tau = \sqrt{\bar{\alpha}_\tau}x_0 + \sqrt{1 - \bar{\alpha}_\tau}\epsilon \qquad
    \epsilon \sim \mathcal{N}\left(0, \mathbf{I}\right)
\end{equation}
Typically, diffusion models are trained to predict the original noise $\epsilon$ given the noisy sample $x_\tau$. We empirically find a different method, known as the $v$-parameterization~\cite{salimans2022progressivedistillation}, to work better in our scenario, where $v$ is defined as:
\begin{align}
    v &= \sqrt{\bar{\alpha}_\tau} \epsilon - \sqrt{1 - \bar{\alpha}_\tau} x_0
\end{align}
This has two advantages. First, $v$-prediction ensures that the loss can always guide the model to learn something meaningful, even when the input already contains a lot of noise, leading to faster convergence~\cite{ho2022imagenvideo}. Second, it allows us to train the model also on pure noise inputs, which during inference is the most challenging denoising step. To this end, we rescale the noise schedule $\alpha$ to ensure zero signal-to-noise ratio inputs during training as proposed by~\cite{lin2023diffusionflawed}.

\subsection{NPHM}
\label{sec:NPHM}

NPHM~\cite{giebenhain2023nphm} is a morphable head model that generates a signed distance field (SDF) of a human head given an identity code $z_{id}$ and an expression code $z_{exp}$. We employ COLMAP~\cite{schoenberger2016mvs_colmap, schoenberger2016sfm_colmap} to obtain pointclouds $\{\mathcal{P}_t\}$ for each timestep $t$ of a multi-view video sequence of the person. Subsequently, we fit NPHM to each pointcloud $\mathcal{P}_t$ to obtain $z_{id}$ and $z^t_{exp}$. We use a variant of NPHM, namely MonoNPHM~\cite{giebenhain2024mononphm}, that uses a backward deformation field instead of the originally proposed forward deformation field as it simplifies the fitting process. Formally, we obtain the NPHM fitting as follows:
\begin{equation}
    z_{id}, z^t_{exp} = \argmin_{z_{id}, z_{exp}} \sum_{x \in \mathcal{P}_t} \left\vert  \mathcal{F}_{id}(\mathcal{F}_{exp}(x)) \right\vert
\end{equation}
where $\mathcal{F}_{id}$ is NPHM's identity network implemented as a signed distance field, and $\mathcal{F}_{exp}$ is the backward deformation network. Note that we drop the dependency of $\mathcal{F}_{id}$ and $\mathcal{F}_{exp}$ on their respective latent codes for simplicity. 
The fitted SDF representation can then be translated into a mesh $M_t$ via marching cubes~\cite{lorensen1998marchingcubes}. Furthermore, for each vertex $x \in M_t$ of the extracted mesh, we can obtain its canonical coordinates $x_{can} \in \mathbb{R}^{3+2}$ via the backward deformation field:
\begin{equation}
    x_{can} = \mathcal{F}_{exp}(x)
\end{equation}
where the first 3 coordinates of $x_{can}$ represent the usual spatial dimensions while the remaining 2 coordinates are ambient dimensions that can help resolve topological issues in mouth and eye regions~\cite{park2021hypernerf}.
We utilize the meshes ${M_t}$ in the subsequent rasterization stage, whereas the canonical coordinates $x_{can}$ are necessary for the spatial feature lookup.

\subsection{Rasterization}
\label{sec:Rasterization}

We employ nvdiffrast~\cite{Laine2020nvdiffrast} to generate the actual input images for the diffusion model. Let $M_t$ be a facial mesh of a person at timestep $t$ and $\pi_t$ a camera pose. We then obtain a set of renderings $R^t \in \mathbb{R}^{H \times W \times C}$ for this timestep using the rasterizer $\mathcal{R}$: 
\begin{align}
    R^t = \mathcal{R}\left(M_t, \pi_t\right)
\end{align}
In our case, the channels of $R^t$ are normals, depths, and a canonical coordinate rendering $R^t_{can} \in \mathbb{R}^{H \times W \times (3 + 2)}$ of the NPHM mesh.

\subsection{TriPlane Feature Mapping}
\label{sec:Feature_Mapping}

It has been shown that tying learnable features to a surface helps neural renderers to synthesize more detailed images, a technique known as neural textures~\cite{thies2019dnr}. Since NPHM lacks a consistent UV space due to its implicit nature, we propose a simple extension of neural textures: We tie learnable features to the surface of the extracted mesh $M_t$ by querying a spatial structure with the canonical coordinates $x_{can}$. We use TriPlanes~\cite{chan2022eg3d} for the 3 spatial dimensions and a regular 2D feature map for the ambient dimensions:
\begin{align}
    R^t_{\text{feat}} &= \textsc{TriPlane}\left(R^t_{can,0\text{-}3}\right) \\
    R^t_{\text{feat\_amb}} &= \textsc{AmbientMap}\left(R^t_{can,3\text{-}5}\right)
\end{align}
These learnable feature maps form the 2D input for our diffusion-based neural renderer, together with the other buffers $R^t$ from the rasterizer:
\begin{align}
    R^t \leftarrow R^t, R^t_{\text{feat}}, R^t_{\text{feat\_amb}}
\end{align}

\subsection{Direct Expression Conditioning}
\label{sec:Expression_Conditioning}

Conditioning on renderings guides the head pose as well as coarse expressions. However, subtle details may not be easy for the network to decode from renderings. To facilitate the synthesis of detailed expressions, we add new cross-attention layers to the U-Net, following IPAdapter~\cite{ye2023ipadapter}. Let $Z$ be the intermediate feature map computed by an existing cross-attention operation in the pre-trained LDM: $Z =\textsc{attention}(Q, K, V)$. We then perform direct expression conditioning by adding another cross-attention layer:
\begin{align}
    f^t_{exp} &= \textsc{exp}(z^t_{exp}) \\
    Z &\leftarrow Z + \textsc{attention}(Q, W^k f^t_{exp}, W^v f^t_{exp})
\end{align}
where \textsc{exp} is our expression conditioning module that linearly maps $z^t_{exp}$ into a sequence $f^t_{exp} \in \mathbb{R}^{4\times d}$ of 4 expression tokens forming the keys and values for cross-attention. In total, we add 15 cross-attention layers to the LDM.

\subsection{Deferred Diffusion}
\label{sec:Deferred Diffusion}

The final rendering is obtained by iteratively denoising full noise $x_{T} \sim \mathcal{N}(0, I)$ with our diffusion-based neural renderer $\mathcal{D}$ conditioned on the rasterized NPHM meshes $R^t$ and expression tokens $f_{exp}$:
\begin{align}
    x_{\tau - 1} \sim \mathcal{D}(x_\tau) = \mathcal{S}(x_\tau, f^t_{exp}, \mathcal{C}(x_\tau, R^t))
\end{align}
where $\mathcal{C}$ is the ControlNet architecture and $\mathcal{S}$ a pre-trained LDM with our expression conditioning modules inserted. The denoised prediction $x_{\tau - 1}$ is obtained via the sampling procedure of~\cite{ho2020ddpm}.

During training, we minimize the following loss:
\begin{align}
    \mathcal{L} = \mathbb{E}_{\epsilon, \tau, x^t_0, R^t, f^t_{exp}}\left[\left\Vert \mathcal{D}\left(x^t_\tau, R^t, f^t_{exp}\right) - v\right\Vert_2\right]
\end{align}
where $\epsilon$ is the sampled noise, $\tau$ is the denoising step and $x^t_0 = \mathcal{E}(I^t)$ is the ground truth latent image. We use $\mathcal{L}$ to optimize (i) the ControlNet $\mathcal{C}$, (ii) the parameters of expression conditioning $W^k, W^v$, and \textsc{exp}, and (iii) the spatial feature maps \textsc{TriPlane} and \textsc{AmbientMap}.

%% file: sections/4_results.tex
\begin{figure*}[htb]
    \centering
    \includegraphics[width=\textwidth]{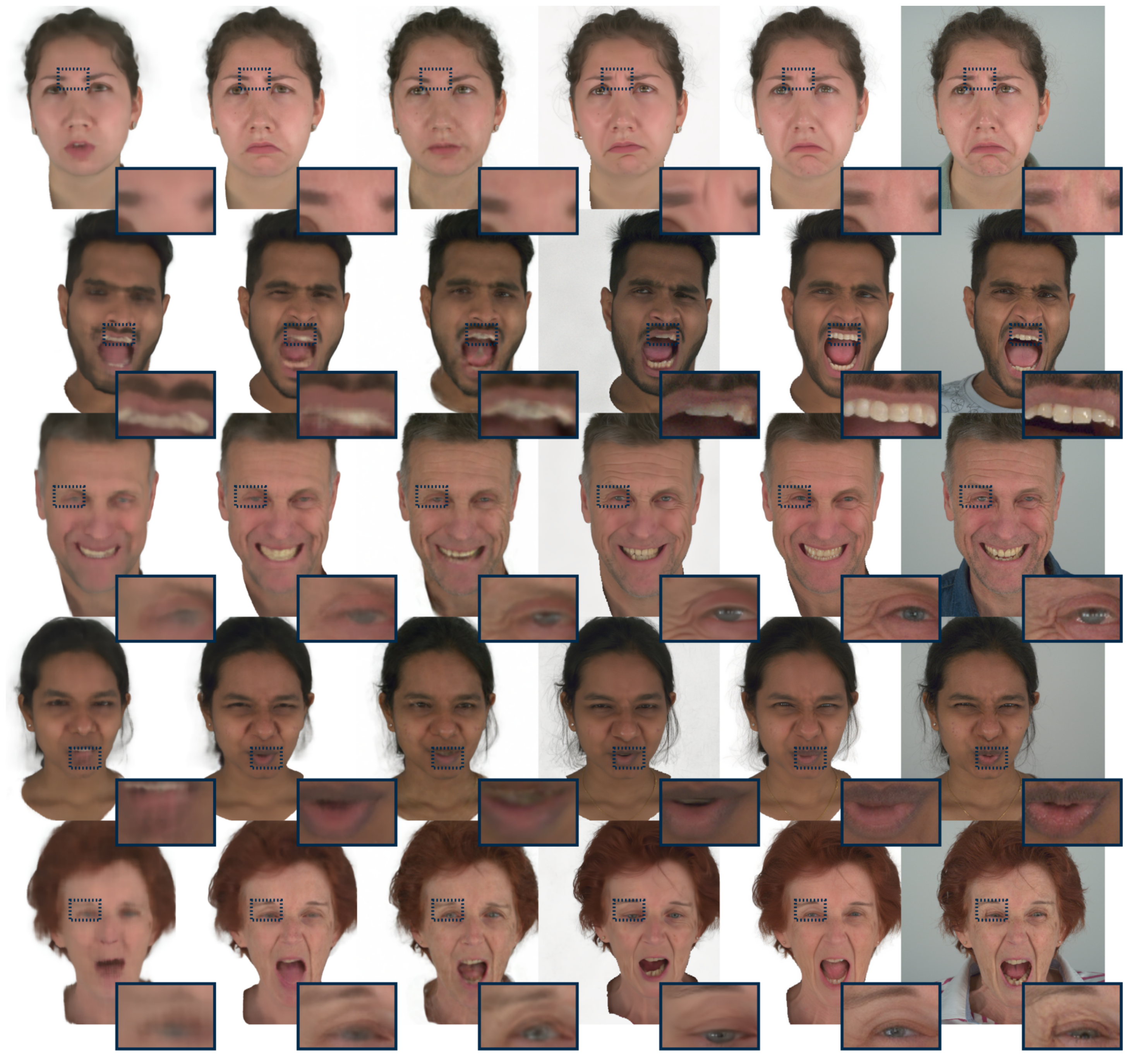}
    \begin{tabularx}{0.95\linewidth}{YYYYYY}
        NeRFace~\cite{gafni2021nerface} & DNR~\cite{thies2019dnr} & MVP~\cite{lombardi2021mixture} & DiffusionRig~\cite{ding2023diffusionrig} & Ours & GT
    \end{tabularx}
    \caption{\textbf{Qualitative results for self-reenactment.} We compare against 3D methods~(NeRFace~\cite{gafni2021nerface}, Mixture of Volumetric Primitives~\cite{lombardi2021mixture}) and methods employing 2D renderers~(Deferred Neural Rendering~\cite{thies2019dnr}, DiffusionRig~\cite{ding2023diffusionrig}). Note that the slightly gray background for DiffusionRig is caused by their training scheme using $\epsilon$-prediction~(see \cref{sec:Diffusion}). Our method consistently produces more expressive facial performances while simultaneously providing more detailed renderings.}
    \label{fig:self-reenactment}
\end{figure*}

\begin{figure*}[htb]
    \centering
    \includegraphics[width=\textwidth]{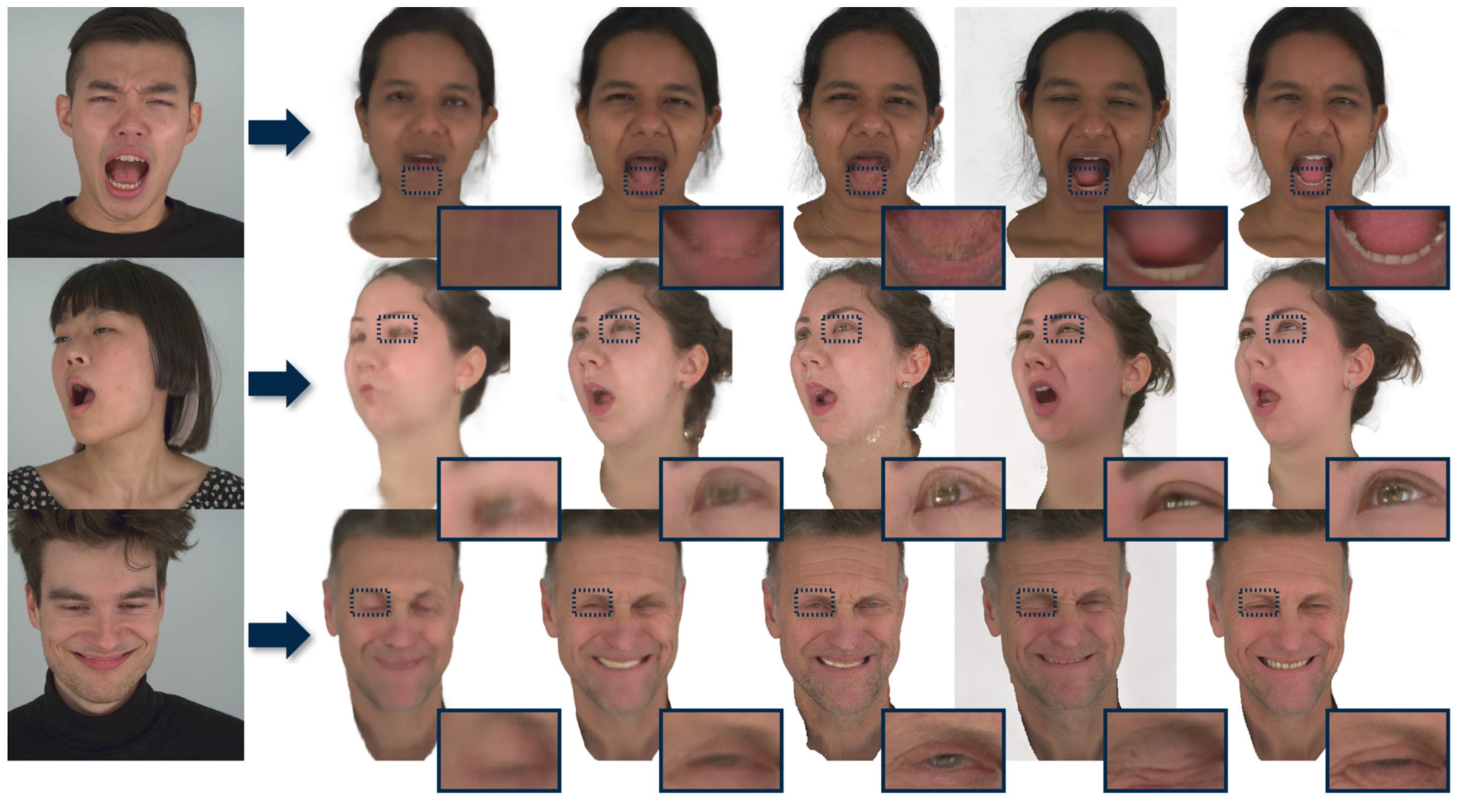}
    \begin{tabularx}{\linewidth}{Yp{0.5pt}YYYYY}
        Source Actor && NeRFace~\cite{gafni2021nerface} & DNR~\cite{thies2019dnr} & DNR + GAN & DiffusionRig~\cite{ding2023diffusionrig} & Ours
    \end{tabularx}
    \caption{\textbf{Qualitative results for avatar animation.} Our method faithfully transfers the source actor's expression and consistently produces compelling renderings, even for complex performances.}
    \label{fig:avatar-animation}
\end{figure*}

\section{Experimental Results}

\subsection{Training}
For our experiments, we employ the recently released multi-view video dataset from NeRSemble~\cite{kirschstein2023nersemble}. The dataset contains 26 different multi-view sequences per person captured with 16 cameras, featuring different facial expressions, sentences and emotions.
We use BackgroundMattingV2~\cite{lin2021backgroundmatting} to remove the background and a segmentation network to remove the torso~\cite{zheng2021farl}. 
We fit NPHM on the video sequences of 8 different persons to create pairings of multi-view videos and fitted meshes.
In total, we use about 3300 timesteps per person. We train one DiffusionAvatar for each person on these pairings with Adam~\cite{kingma2014adam} using a learning rate of $1e-4$ for ControlNet and the expression conditioning layers, and a learning rate of $1e-2$ for the TriPlane feature lookups. We use a batch size of $B=8$ and spatial resolutions $H = W = 512$. We employ random cropping and resizing of the target images to avoid overfitting. The size of our spatial feature maps is $512\times512\times16$ for each TriPlane and the ambient map. In total, this yields $16 \times (3+1) + 5 + 1 + 3 = 73$ channels for the input renderings $R_t$ fed to our diffusion-based renderer. The pre-trained LDM is a Stable Diffusion v2.1 network~\cite{rombach2022ldm}. Training takes roughly two days on a single RTX A6000 for 100k gradient descent steps.

\input{tables/04_cr_results_self_reenactment}

\subsection{Experiment Setup}

\paragraph{Tasks.}
To measure the quality of a 3D avatar, we conduct experiments on two tasks: 
Self-reenactment and avatar animation. For self-reenactment, we hold out 2 complete sequences and the frontal camera of every sequence during training to measure both image quality and view consistency. During testing, we provide the holdout fitted meshes of the same person and render from the novel view. For avatar animation, we first generate NPHM meshes using the same identity code $z_{id}$ as the avatar and a sequence of expression codes $\{z^t_{exp}\}$ from a different person. We then render a video of the avatar from a novel view. To evaluate the realism of this expression transfer, we conduct a user study where we show the animated avatars to participants.
\paragraph{Metrics.}
We employ two paired-image metrics to measure the quality of individual generated frames: Peak Signal-to-Noise Ratio (PSNR) and Learned Perceptual Image Patch Similarity (LPIPS) \cite{zhang2018lpips}. Furthermore, we measure visual similarity of a rendered video to its ground truth counterpart with the perceptual video metric JOD~\cite{jod_metric}. Finally, we make use of several face-specific metrics: Average Keypoint Distance (AKD) measured in pixels with keypoints estimated from PIPNet~\cite{jin2021pipnet}, and cosine similarity (CSIM) of identity embeddings based on ArcFace~\cite{deng2019arcface}. We further follow~\cite{chu2024gpavatar} and utilize DECA~\cite{DECA:Siggraph2021} to measure Average Expression Distance (AED) and Average Pose Distance (APD) between the rendered and the ground truth images.

\paragraph{Baselines.}
We compare our method against the following state-of-the-art 3D head avatar creation methods: 

\textit{NeRFace}~\cite{gafni2021nerface}. A NeRF-based method that reconstructs a radiance field conditioned on expression codes. The original implementation is for monocular videos and uses FLAME. We extend it to our multi-view training scenario and provide NPHM expression codes for better comparison.

\textit{Mixture of Volumetric Primitives} (MVP)~\cite{lombardi2021mixture}. An encoder-decoder architecture that generates a set of small feature grids rigged to a template mesh. We provide a FLAME tracking to learn the driving.

\textit{Deferred Neural Rendering} (DNR)~\cite{thies2019dnr}. A screen-space decoder built on top of Pix2Pix~\cite{zhu2017pix2pix} that uses learnable neural textures rigged to a tracked Basel Face Model~\cite{paysan2009bfm}. We provide our fitted NPHM meshes as input for better comparison. Additionally, we compare against a version of DNR that also uses an adversarial loss for sharper images that we call DNR+GAN in the following.

\textit{DiffusionRig}~\cite{ding2023diffusionrig}. A recent diffusion-based method that uses albedo, normal, and shaded color renderings of FLAME meshes tracked by DECA~\cite{DECA:Siggraph2021} as input to control pose and expression. While their method was originally proposed to be fine-tuned on a small personal photo album of ~20 images, we fine-tune their pre-trained model on all training frames of our dataset.

\subsection{Self-Reenactment}
In~\cref{tab:results_self_reenactment}, we compare {\OURS} to other state-of-the-art approaches on the self-reenactment task. Results are averaged over avatars generated from eight different persons. Our method shows the overall best performance with the biggest improvement in perceptual and face-specific metrics. The results indicate that {\OURS} generates sharper images~(LPIPS) that are more temporally consistent~(JOD). Furthermore, our method better maintains the person's identity when rendered from a novel view~(CSIM) than the baselines. Finally, facial features such as the position of mouth corners and eyebrows synthesized by our method match the ground truth expressions more closely~(AKD, AED, APD).\\
We further provide a qualitative comparison in~\cref{fig:self-reenactment}. The baselines struggle especially in facial regions that are poorly explained by the NPHM mesh, such as the mouth interior. In contrast, our method can plausibly fill these regions, providing a consistently realistic appearance.

\input{tables/04_cr_results_avatar_animation}

\subsection{Avatar Animation}
Due to the task's inherent lack of any ground truth,  we perform a user study to validate avatar animation. We collect {\nresponses} responses from {\nparticipants} participants on two questions:

(i)~Visual Quality~(VQ): What is the overall quality of the avatar animation?

(ii)~Driving Fidelity~(DF): How closely does the animated avatar match the expressions of the driving sequence?

We present two videos side-by-side: One sequence of a source actor performing various expressions, and a video of another person's avatar with the expressions transferred to it. The results of our user study can be seen in~\cref{tab:results_avatar_animation}. Our method achieves the highest scores for both visual quality and similarity to the driving sequence. The difference is particularly pronounced for the visual quality score. We attribute this to the fact that users heavily penalize any visual artifacts immediately obvious in video renderings.
These perceptual differences in the task of avatar animation are also apparent in the qualitative comparison shown in~\cref{fig:avatar-animation}.

\subsection{Ablations}

In~\cref{tab:ablation}, we examine the design choices of our method. All ablation experiments are conducted on 3 avatars on the self-reenactment task. We refer to our supplementary material for a qualitative comparison.

\input{tables/04_ablation}

\paragraph{Effect of Diffusion.} 
To study whether diffusion is necessary, we train our architecture without the pre-trained LDM to directly predict the output in a single forward pass, similar to Deferred Neural Rendering. Such a method performs notably worse, especially in the LPIPS metric, indicating that diffusion is important for sharp renderings.

\paragraph{Effect of NPHM vs FLAME.}
We compare {\OURS} with a version trained on renderings of a fitted FLAME instead of NPHM. 
Since NPHM models facial geometry as an SDF, it possesses a much higher representational power than FLAME. 
As a result, the renderings are much closer to the actual desired appearance of the head. This is also reflected in our ablation study, where using FLAME gives consistently worse performance. This is especially true for the AKD metric, indicating that our diffusion-based neural renderer cannot reenact certain complex expressions based on FLAME alone.

\paragraph{Effect of Expression Conditioning.}

Turning off expression conditioning worsens overall performance, 
presumably because the expression codes help to distinguish subtle expression details that can hardly be inferred from the NPHM mesh alone. It also helps {\OURS} to synthesize expressions not captured by NPHM, such as tongue movement, by enabling the network to correlate such appearances with certain expression code combinations.

\paragraph{Effect of 2D LDM Prior.}
We train our pipeline from scratch without resorting to a pre-trained LDM. In this case, we remove the ControlNet module and condition the U-Net directly on the NPHM renderings. This architecture, trained from scratch, performs competitively but slightly worse. 
We see the biggest advantage of pre-training in faster convergence and better coherence of the generated images.

\paragraph{Effect of Spatial Features.} 
We compare three spatial feature mapping approaches: (i) Our Triplanes, (ii) a simpler approach based on spherical feature mapping, and (iii) no spatial features at all.
For spherical feature mapping, we create an approximate UV map for each NPHM mesh by projecting the 3D canonical coordinates for each vertex onto a sphere centered inside the head. \cref{tab:ablation} shows that using TriPlanes boosts most metrics, whereas the simpler spherical mapping brings almost no benefit.

\section{Limitations}

{\OURS} can create photo-realistic 3D head avatars with pose and expression control. More challenges must be solved, however, before an avatar can be used in production. For example, to put an avatar into a realistic environment, it is necessary to have control over the lighting properties. Currently, {\OURS} bakes the lighting into the generated images. Since we employ a detailed 3D head geometry underneath, one could imagine directly modeling shadow effects as a composite of the synthesized images. Further, our current architecture is not (yet) amenable to real-time applications due to the comparatively compute-intensive denoising loop. Here, recent advances in the distillation of diffusion models may provide a remedy~\cite{song2023consistency, meng2023distillation}.

\section{Conclusion}
This paper presents a novel method {\OURS}, which creates a photo-realistic 3D head avatar from multi-view videos using a tracked parametric head model of a person. DiffusionAvatars can be animated by a source actor or directly via NPHM. We use rendered NPHM meshes with rigged spatial features as input to a diffusion-based neural renderer. We leverage the image synthesis capabilities of a pre-trained latent diffusion model with ControlNet, facilitating generalization to unseen expressions. 
While this pipeline is already capable of synthesizing appealing images, we show that additionally conditioning the neural renderer on NPHM's expression codes further improves the model's ability to generate complex facial expressions. Our experiments suggest that the proposed architecture can effectively synthesize high-quality renderings and animate the 3D avatar with high fidelity. We believe our approach demonstrates an exciting application for high-fidelity 3D avatar creation.

%% file: tables/04_cr_results_self_reenactment.tex
\begin{table}[htb]
    
    \setlength{\tabcolsep}{2pt}
    \centering
    \resizebox{\linewidth}{!}{%
        \begin{tabular}{lrrrrrrrr}
            \toprule
            Method
            & \footnotesize{PSNR}$\uparrow$ & \footnotesize{LPIPS}$\downarrow$ & \footnotesize{JOD}$\uparrow$ & \footnotesize{AKD}$\downarrow$ & \footnotesize{AED}$\downarrow$ & \footnotesize{APD}$\downarrow$ & \footnotesize{CSIM}$\uparrow$ \\
            
            \midrule
            
            NeRFace~\cite{gafni2021nerface} %
                & 23.0 & 0.279 & 6.76 &	5.37 & 1.06 & 0.053 & 0.787 \\
            DiffusionRig~\cite{ding2023diffusionrig}
                & 19.6 & 0.220 & 6.41 & 2.74 & 0.55 & 0.029 & 0.887 \\
            DNR~\cite{thies2019dnr} 
                & 24.5 & 0.226 & 7.32 & 2.06 & 0.63 & 0.027 & 0.903 \\
            DNR+GAN~\cite{thies2019dnr}
                & 23.0 & 0.114 & 7.08 & 2.14 & 0.69 & 0.028 & 0.868 \\
            MVP~\cite{lombardi2021mixture}
                & 23.6 & 0.221 & 7.02 & 3.42 & 0.78 & 0.034 & 0.882 \\
            Ours
                & \textbf{24.9} & \textbf{0.081} & \textbf{7.55} & \textbf{1.79} & \textbf{0.50} & \textbf{0.023} & \textbf{0.917} \\
    
            \bottomrule
            
        \end{tabular}%
    }
    
    \caption{\textbf{Quantitative Comparison:} We report metrics for a self-reenactment scenario on unseen expressions and unseen views averaged over 8 persons from the NeRSemble dataset~\cite{kirschstein2023nersemble}.}

    \label{tab:results_self_reenactment}
    
\end{table}

%% file: tables/04_cr_results_avatar_animation.tex
\begin{table}[htb]
    
    \setlength{\tabcolsep}{3.5pt}
    \centering
    \resizebox{\linewidth}{!}{%
        \begin{tabularx}{\linewidth}{lccccc}
            \toprule
            & NeRFace & DiffusionRig & DNR  & DNR+GAN & Ours\\

            \midrule
            
            VQ$\uparrow$ & 2.19 & 2.47 & 2.87 & 3.06 & \textbf{4.02} \\
            DF$\uparrow$ & 2.35 & 3.52 & 3.94 & 3.97 & \textbf{4.14} \\

            \bottomrule
            
        \end{tabularx}%
    }
    
    \caption{\textbf{User Study:} For our user study on avatar animation, we collect {\nresponses} responses from {\nparticipants} participants to measure Visual Quality (VQ) and Driving Fidelity (DF) on a scale of 1-5.}

    \label{tab:results_avatar_animation}
    
\end{table}

%% file: tables/04_ablation.tex
\newcommand{\checkbox}{\scalebox{1.5}{$\boxtimes$}}
\newcommand{\emptybox}{\scalebox{1.5}{$\square$}}
\newcommand{\sphericalbox}{\scalebox{1.5}{S}}
\newcommand{\triplanebox}{\scalebox{1.5}{T}}

\begin{table}[htb]
    \setlength{\tabcolsep}{3.5pt}
    \centering
    
    \begin{tabular}{lrrrrr}
        \toprule

        & \footnotesize{PSNR}$\uparrow$ & \footnotesize{LPIPS}$\downarrow$ & \footnotesize{JOD}$\uparrow$ & \footnotesize{AKD}$\downarrow$ & \footnotesize{CSIM}$\uparrow$ \\

        \midrule

        w/o diffusion    & 25.1 & 0.133 & 7.69 & 1.96 & 0.892 \\
        FLAME 			& 24.2 & 0.083 & 7.36 & 2.46 & 0.900 \\
        w/o exp. cond. 	& 24.5 & 0.081 & 7.65 & 1.98 & 0.911 \\
        w/o LDM prior 	& 24.5 & 0.078 & 7.67 & 1.89 & 0.913 \\
        w/o spatial features		& 24.9 & 0.078 & 7.77 & \textbf{1.87} & 0.918 \\
        spherical UV 	& 24.9 & 0.075 & 7.77 & 1.95 & \textbf{0.920} \\
        Ours            & \textbf{25.3} & \textbf{0.074} & \textbf{7.85} & 1.91 & 0.918 \\

        \bottomrule
    \end{tabular}
    \caption{\textbf{Ablation of architectural choices.} 
    }
    \vspace{-0.5cm}
    \label{tab:ablation}
\end{table}

%% file: sections/5_acknowledgements.tex
\subsection*{Acknowledgements}
This work was supported by the ERC Starting Grant Scan2CAD (804724) and the German Research Foundation (DFG) Research Unit ``Learning and Simulation in Visual Computing''.
We would also like to thank Karla Weighart for proof-reading and Angela Dai for the video voice-over.

%% file: sections/X_suppl.tex
\clearpage

\appendix
\bigskip

\huge \textbf{Appendix}
\normalsize

\section{Additional Results}

\begin{figure*}[htb]
    \centering
    \includegraphics[width=\textwidth]{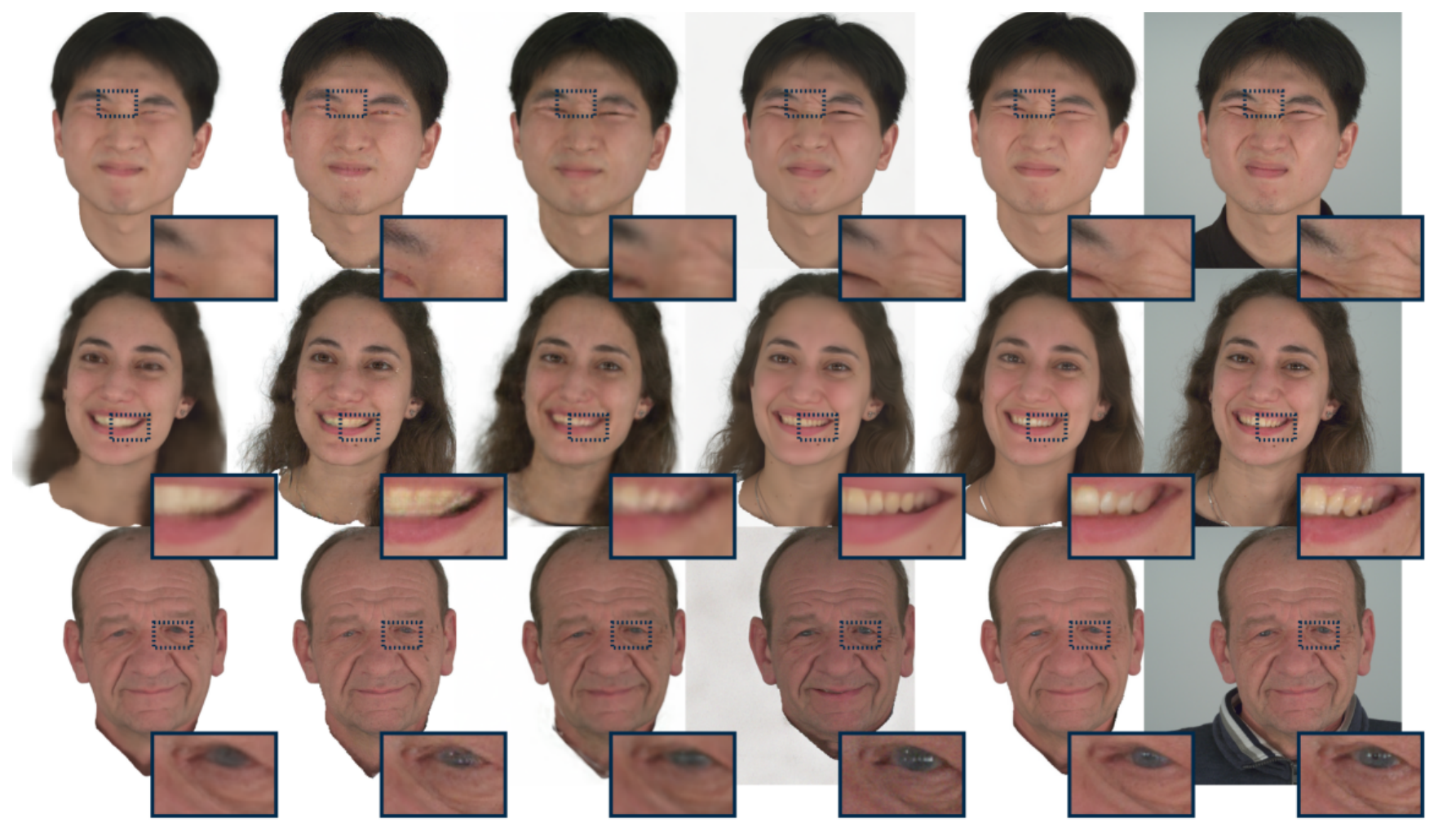}
    \begin{tabularx}{0.95\linewidth}{YYYYYY}
        DNR~\cite{thies2019dnr} & DNR+GAN & MVP~\cite{lombardi2021mixture} & DiffusionRig~\cite{ding2023diffusionrig} & Ours & GT
    \end{tabularx}
    \caption{\textbf{Qualitative results for self-reenactment.}}
    \label{fig:supp-self-reenactment}
\end{figure*}

We provide additional qualitative results for the self-reenactment task on three more people from the NeRSemble~\cite{kirschstein2023nersemble} dataset in~\cref{fig:supp-self-reenactment}.

\subsection{Novel View Synthesis Results}

\cref{tab:results_novel_view_synthesis} contains results for novel view synthesis. We measure the performance from the unseen frontal view and average over the 24 train sequences for each of our 5 avatars. {\OURS} scores the best metrics except for PSNR, which favors the overly blurry renderings of NeRFace. These results show that our method also provides good 3D control.

\input{tables/X_results_novel_view_synthesis}

\subsection{Ablation of Number of Training Views}

\begin{figure}[htb]
    \centering
    \includegraphics[width=\linewidth]{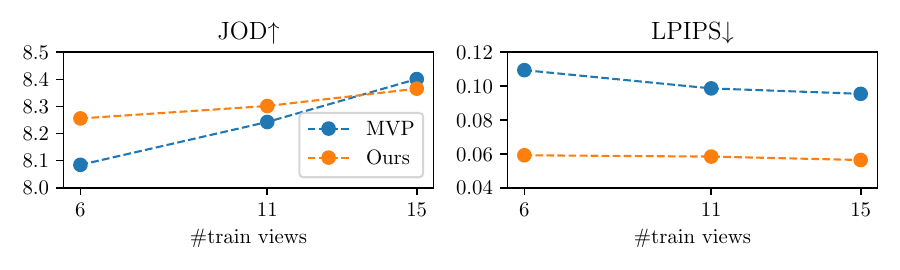}
    \caption{Ablation of number of cameras on the Multiface dataset.}
    \label{fig:rebuttal_n_cameras_ablation}
\end{figure}

We analyze the data efficiency of DiffusionAvatars on one person of the Multiface dataset~\cite{wuu2022multiface} in~\cref{fig:rebuttal_n_cameras_ablation}. Our method only suffers negligibly from the removal of train views, as opposed to MVP.
We attribute this to NPHM's strong prior over 3D head shapes which allows fitting to sparse pointclouds, while direct 3D methods like MVP struggle more to recover accurate surfaces from sparse views.

\subsection{Qualitative Comparison of Ablations}

\begin{figure}[htb]
    \centering
    \includegraphics[width=\linewidth]{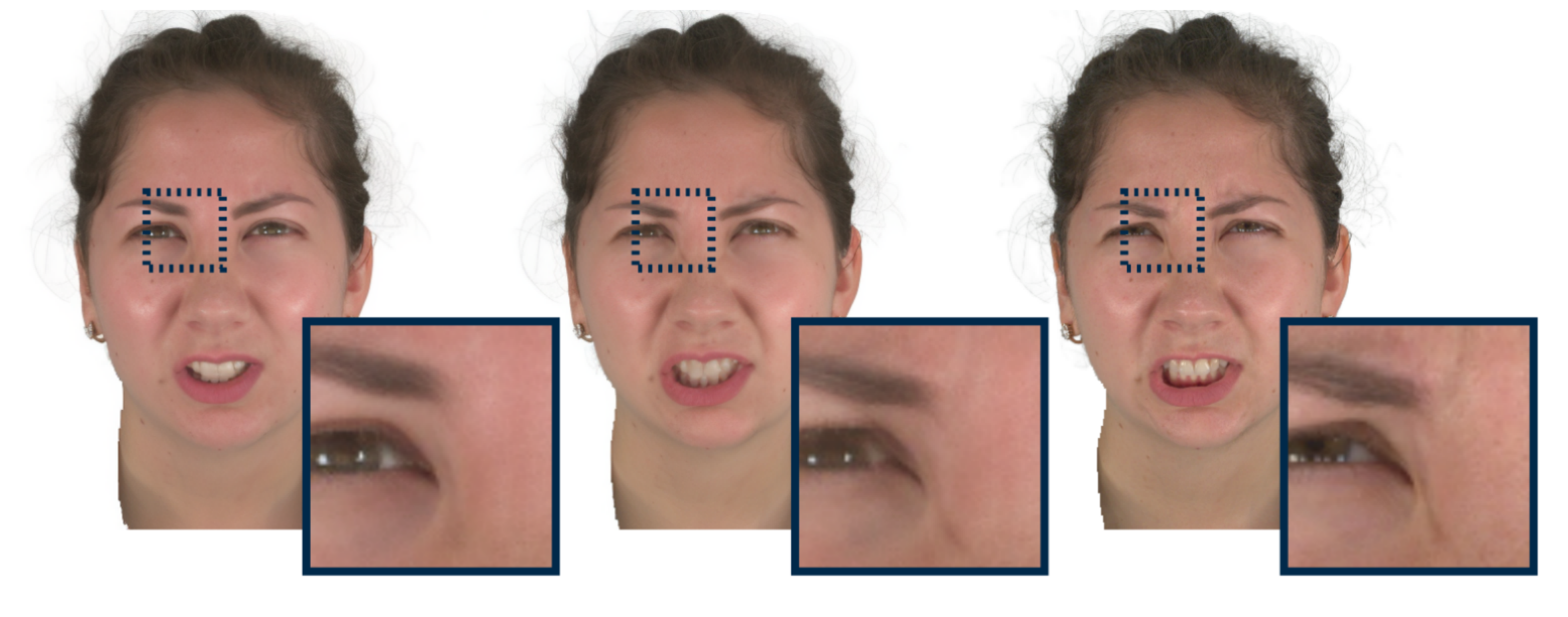}
    \begin{tabularx}{\linewidth}{YYY}
        FLAME~\cite{FLAME:SiggraphAsia2017} & Ours & GT
    \end{tabularx}
    \caption{\textbf{Effect of FLAME vs. NPHM.} Using FLAME instead of NPHM as the underlying morphable head model makes the generated faces less expressive. This is due to FLAME's limited geometry and expression space, which does not provide enough geometric cues for our diffusion-based neural renderer.}
    \label{fig:ablation_FLAME}
\end{figure}

\begin{figure}[htb]
    \centering
    \includegraphics[width=\linewidth]{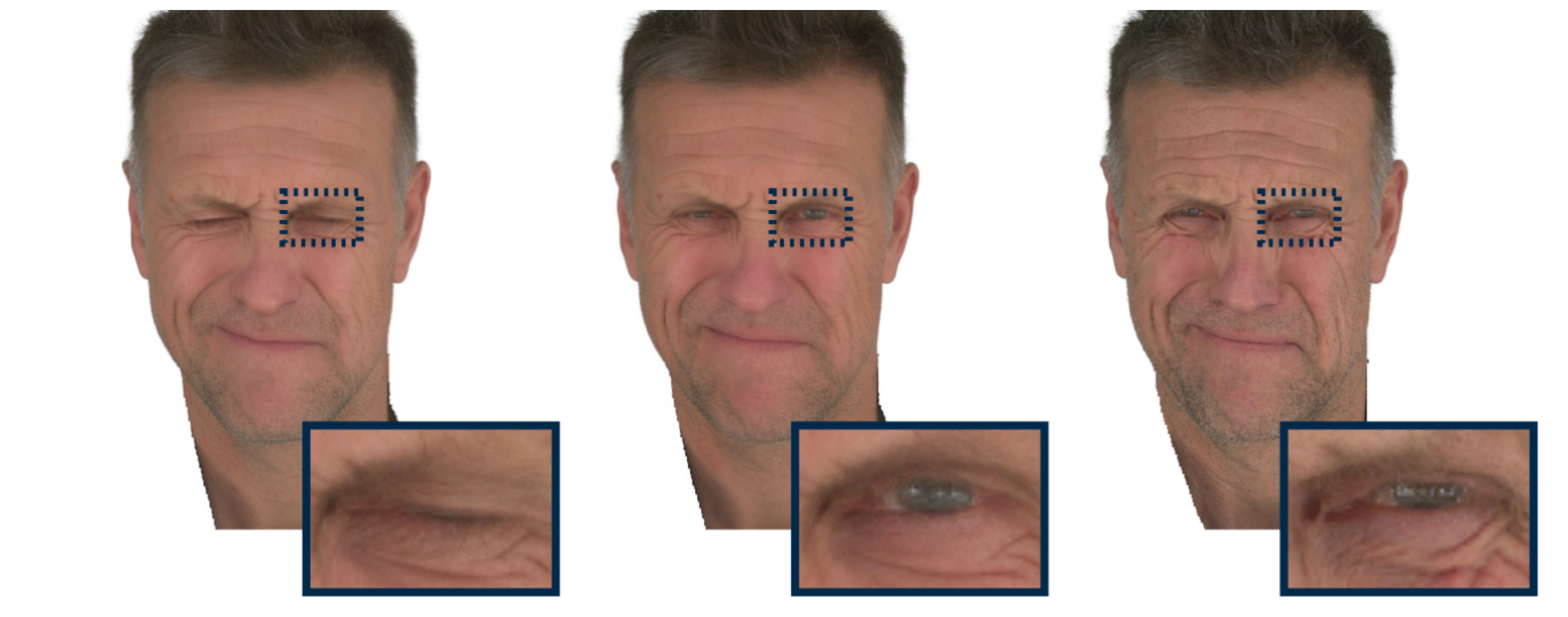}
    \begin{tabularx}{\linewidth}{YYY}
        w/o exp. cond. & Ours & GT
    \end{tabularx}
    \caption{\textbf{Effect of expression conditioning.} Directly providing $z_{exp}$ to the diffusion model helps synthesize areas not modeled by NPHM's geometry, such as eye movement.}
    \label{fig:ablation_wo_exp}
\end{figure}

\begin{figure}[htb]
    \centering
    \includegraphics[width=\linewidth]{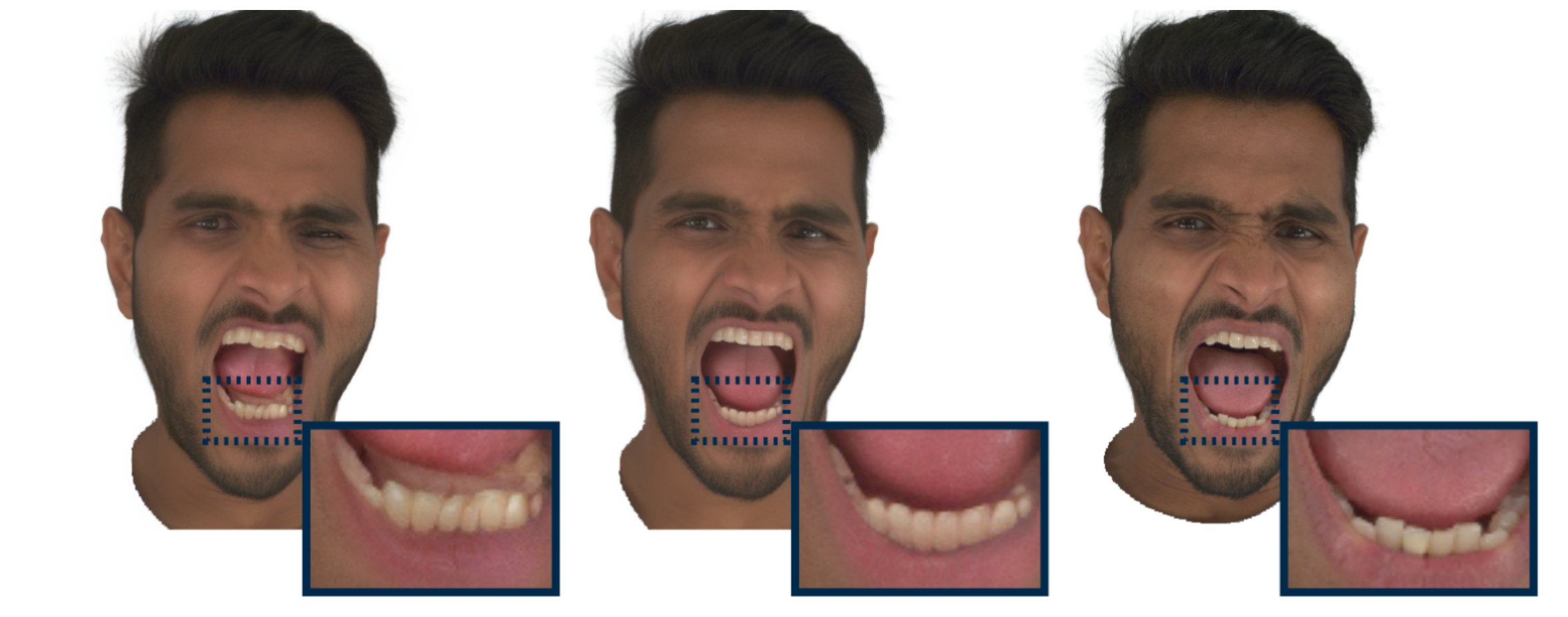}
    \begin{tabularx}{\linewidth}{YYY}
        w/o LDM prior & Ours & GT
    \end{tabularx}
    \caption{\textbf{Effect of LDM prior.} Using a pre-trained LDM instead of training the diffusion model from scratch steers the generation towards more realistic faces.}
    \label{fig:ablation_ldm_prior}
\end{figure}

As an addition to \cref{tab:ablation} of the main paper, we provide qualitative comparisons of three ablations in \cref{fig:ablation_FLAME}, \cref{fig:ablation_wo_exp} and \cref{fig:ablation_ldm_prior}.

\subsection{Experiments on the Multiface dataset}

\begin{figure}[htb]
    \setlength{\tabcolsep}{0pt}
    \centering
    \def\arraystretch{0.5}
    \begin{tabularx}{\textwidth}{
         P{0.33\linewidth}%
         P{0.33\linewidth}%
         P{0.33\linewidth}%
     }

    \multicolumn{3}{c}{\includegraphics[width=\linewidth]{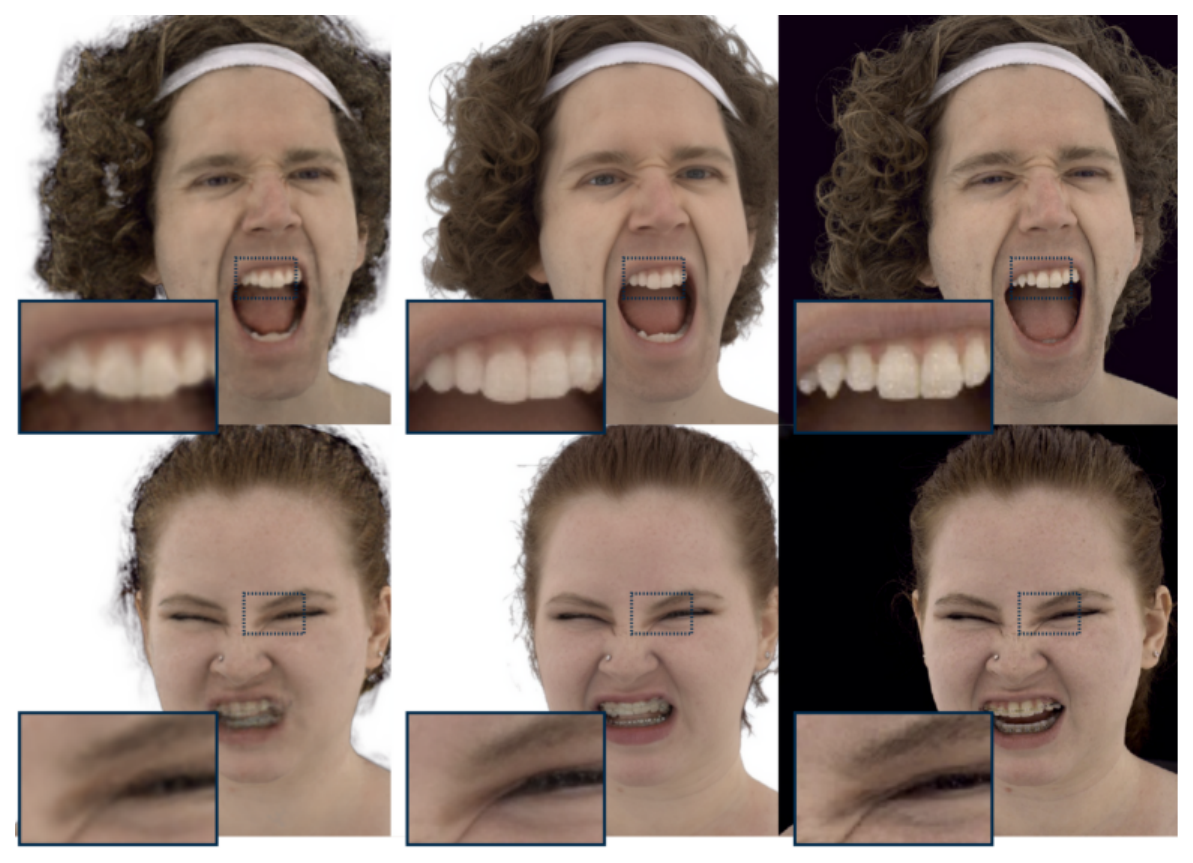}}
    \\
        MVP~\cite{lombardi2021mixture} & Ours & GT
    \end{tabularx}
    \caption{Qualitative comparison for self-reenactment on the Multiface dataset.}
    \label{fig:rebuttal_self_reenactment}
\end{figure}

We conduct additional experiments on 16 cameras from the Multiface dataset~\cite{wuu2022multiface} in~\cref{tab:rebuttal_results_multiface}. Although MVP has the advantage of a high-end textured mesh tracking pre-computed from 100+ cameras there, DiffusionAvatars performs competitively showing that our method (i) can compete with recent 3D methods and (ii) generalizes to different datasets. In~\cref{fig:rebuttal_self_reenactment}, we show that our method produces sharper renderings than MVP on the Multiface dataset.

\input{tables/X_results_multiface}

\section{User Study Setup}

\begin{figure}[htb]
    \centering
    \includegraphics[width=\linewidth]{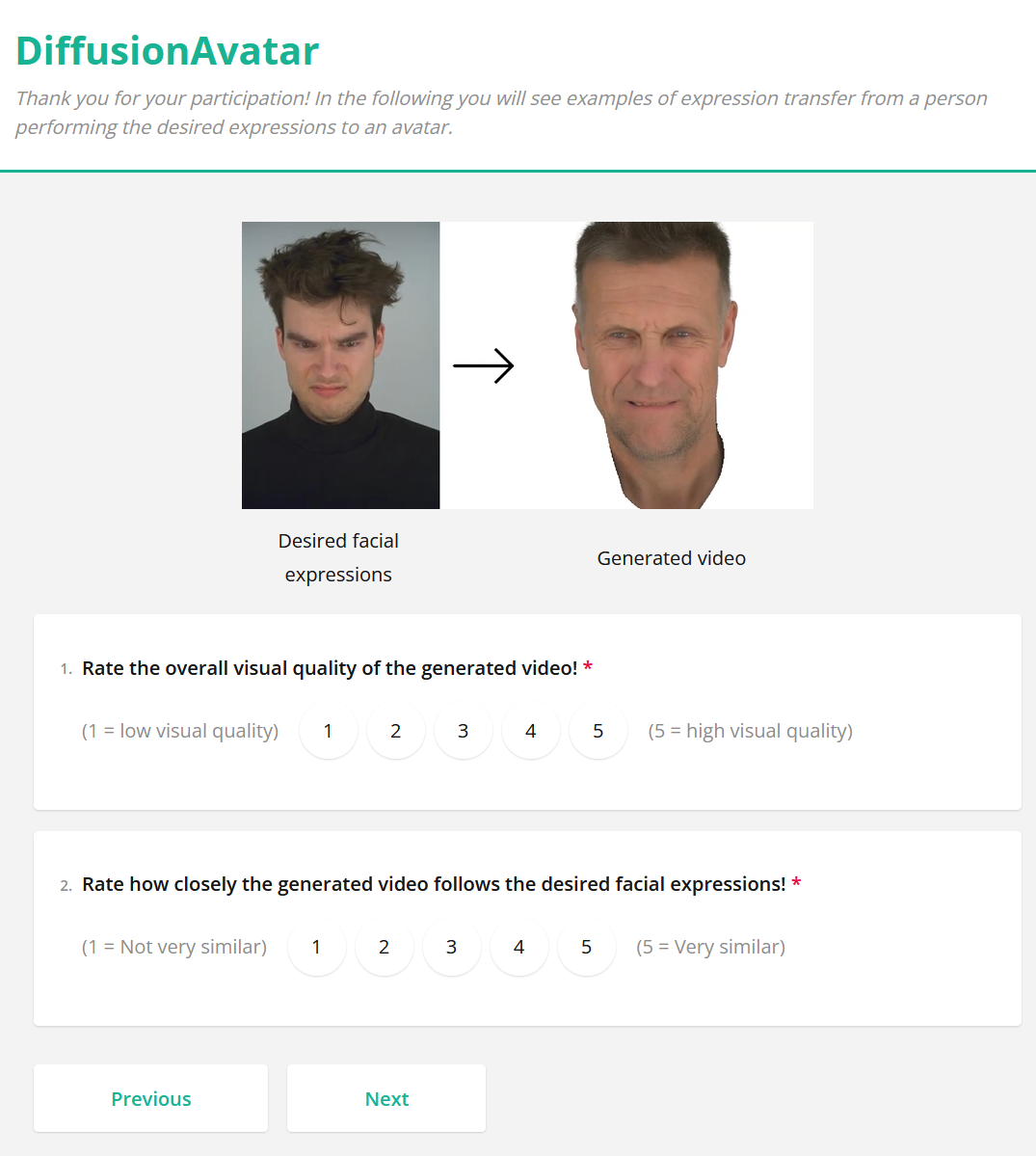}
    \caption{\textbf{Setup of our User Study.} We ask participants to evaluate the avatar animation task in which an avatar is controlled by a sequence of a different person.}
    \label{fig:user_study_setup}
\end{figure}

In ~\cref{fig:user_study_setup}, we show the interface of our user study. We ask users to rate visual quality and driving fidelity by providing them with the following definitions:
\paragraph{Visual Quality.} The overall quality of the generated video: Is there any flickering, visual artifacts, blurriness, etc., that feel unnatural?
\paragraph{Driving Fidelity.} How closely does the generated video follow the desired facial expressions: Are the displayed expressions the same? Is the emotion displayed by the avatar recognizable?
\\\\
During the study, each user is presented with 15 randomly chosen pairs of driving sequences (left) and avatar animations (right). The pairs were drawn from 135 different avatar animations generated by the baselines and our method. In total, we received {\nresponses} responses for each question from {\nparticipants} participants.

\section{Analysis of Temporal Consistency}
\begin{figure*}[htb]
    \setlength{\tabcolsep}{0pt}
    \centering
    \def\arraystretch{0.7}
    \begin{tabularx}{\linewidth}{%
        P{0.24\linewidth}%
         P{0.27\linewidth}%
         P{0.24\linewidth}%
         P{0.27\linewidth}%
     }%
     MVP & DNR+GAN & Ours & GT \\
     \scriptsize{$\leftarrow$ time $\rightarrow$} 
     & \scriptsize{$\leftarrow$ time $\rightarrow$}
     & \scriptsize{$\leftarrow$ time $\rightarrow$}
     & \scriptsize{$\leftarrow$ time $\rightarrow$}\\
     \end{tabularx}
    \includegraphics[width=\textwidth]{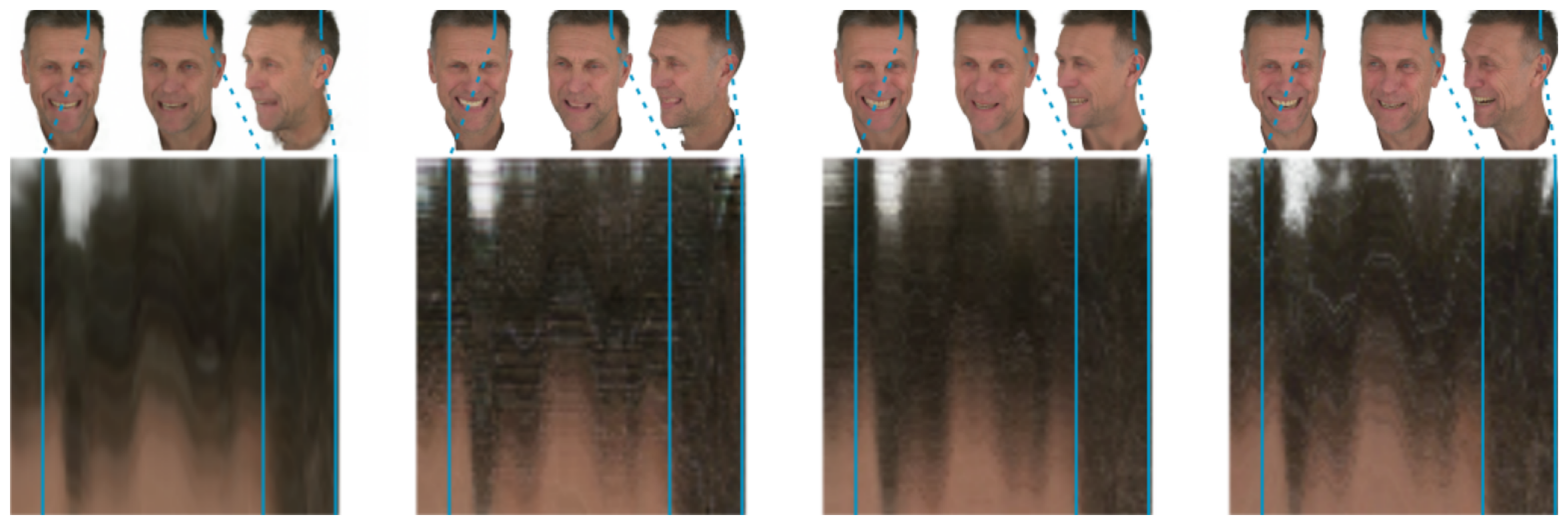}
    \caption{\textbf{Analysis of Temporal Consistency.} We extract a short vertical segment of pixels from each frame of a generated video and concatenate them left-to-right (bottom). The desired result is that texture details move together with the underlying head geometry. It can be clearly seen that combining Deferred Neural Rendering with an adversarial loss~(DNR+GAN) causes severe texture sticking as the person rotates their head to the left. On the other hand, MVP and our method do not suffer from this issue. Additionally, as opposed to MVP, our method generates much of the high-frequency detail that is also present in the ground truth video.  
    }
    \label{fig:temporal_analysis}
\end{figure*}

Since {\OURS} is based on 2D diffusion models, the generated images may be susceptible to view-inconsistencies and screen-space stitching artifacts. Following StyleGAN3~\cite{karras2021stylegan3}, we therefore perform an analysis of the temporal consistency of generated videos. In figure~\cref{fig:temporal_analysis}, we compare our method to MVP~\cite{lombardi2021mixture} and Deferred Neural Rendering~\cite{thies2019dnr}. Our finding is that {\OURS} generates images with more high-frequency detail than 3D methods while being significantly less prone to texture-sticking than other 2D methods. 

\section{Model Architecture}
\begin{figure*}[htb]
    \centering
    \includegraphics[width=\textwidth]{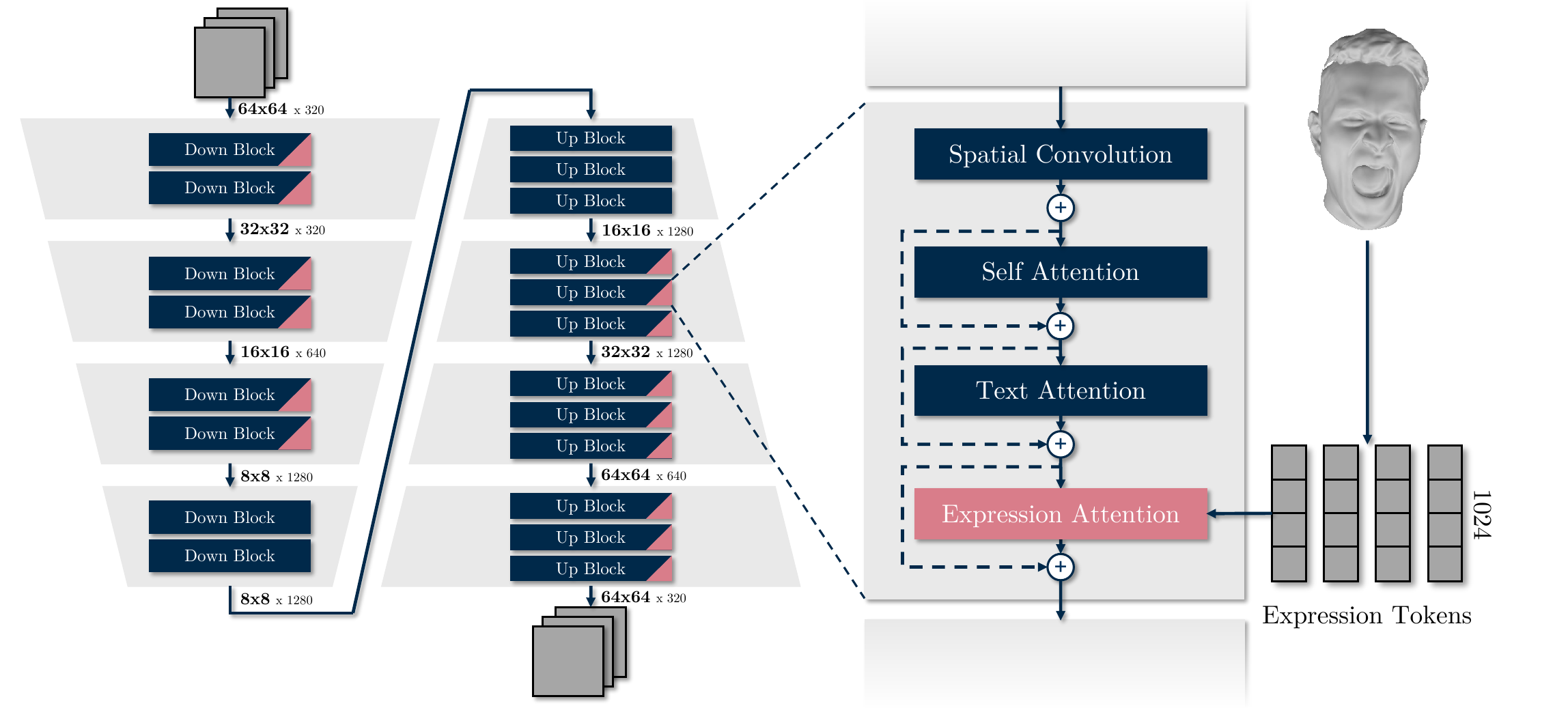}
    \begin{tabularx}{\linewidth}{
        P{0.4\linewidth}
        P{0.5\linewidth}
    }
        (a) Diffusion U-Net & (b) Expression Attention Block
    \end{tabularx}
    \caption{\textbf{Architecture of DiffusionAvatars.} We build on the U-Net architecture of a pre-trained Stable Diffusion~\cite{rombach2022ldm} model. While we keep all original layers fixed, we introduce an additional cross attention layer in each block that allows spatial features to attend to expression tokens extracted from NPHM's expression codes. That way, subtle expression details can influence the image formation process. Additionally, following the ControlNet~\cite{zhang2023controlnet} paradigm, the encoder part of the diffusion U-Net is copied and conditioned on the NPHM rasterizations. The feature maps produced by this ControlNet are added to the existing skip connections of the U-Net, allowing the diffusion model to generate articulated facial images while keeping the knowledge from its pre-training.}
    \label{fig:architecture}
\end{figure*}

\cref{fig:architecture} depicts {\OURS}'s model architecture.

\section{Societal Impact}
DiffusionAvatars provides the means to generate realistic images of faces with control over viewpoint, pose, and expression. As such, an avatar could be misused to the original person's disadvantage. However,
we focus on 3D applications such as immersive teleconferencing or character animation, not indistinguishable 2D video synthesis. Our method does not model lighting, the background, or the torso. Therefore, we believe a detection system can be built to identify generations of our method by exploiting such characteristics~\cite{rossler2018faceforensics}.\\
All subjects shown in the paper and the supplemental video are part of the NeRSemble dataset~\cite{kirschstein2023nersemble} and have consented to their recordings being used for research purposes.

%% file: tables/X_results_novel_view_synthesis.tex
\begin{table}[htb]
    
    \setlength{\tabcolsep}{3pt}
    \centering
    \resizebox{\linewidth}{!}{%
        \begin{tabular}{lrrrrrr}
            \toprule
            Method 
            & \footnotesize{PSNR}$\uparrow$ & \footnotesize{LPIPS}$\downarrow$ & \footnotesize{JOD}$\uparrow$ & \footnotesize{AKD}$\downarrow$ & \footnotesize{CSIM}$\uparrow$ \\
            
            \midrule
            
            NeRFace~\cite{gafni2021nerface} %
                & \textbf{28.12} & 0.284 & 7.93 & 2.72 & 0.841 \\
            DiffusionRig~\cite{ding2023diffusionrig}
                & 19.73 & 0.231 & 6.02 & 9.71 & 0.917 \\
            DNR~\cite{thies2019dnr} 
                & 26.90 & 0.213 & 7.91 & 1.58 & 0.933 \\
            DNR+GAN~\cite{thies2019dnr}
                & 23.56 & 0.104 & 7.30 & 1.70 & 0.906 \\
            Ours
                & 26.93 & \textbf{0.067} & \textbf{8.28} & \textbf{1.16} & \textbf{0.939} \\
    
            \bottomrule
            
        \end{tabular}%
    }
    
    \caption{\textbf{Quantitative Comparison:} We report metrics for novel-view-synthesis averaged over five persons from the NeRSemble dataset.}

    \label{tab:results_novel_view_synthesis}
    
\end{table}

%% file: tables/X_results_multiface.tex
\begin{table}[htb]
    \setlength{\tabcolsep}{3.5pt}
    \centering
    \resizebox{\linewidth}{!}{%
        \begin{tabular}{lrrrrrrr}
            \toprule
                 & \footnotesize{PSNR}$\uparrow$ & \footnotesize{LPIPS}$\downarrow$ & \footnotesize{JOD}$\uparrow$ & \footnotesize{AKD}$\downarrow$ & \footnotesize{AED}$\downarrow$ & \footnotesize{APD}$\downarrow$ & \footnotesize{CSIM}$\uparrow$\\
             \midrule
             MVP
                & \textbf{26.4} & 0.126 & \textbf{7.92} & 2.25 & 0.655 & 0.026 & 0.904 \\
             Ours 
                & 25.5 & \textbf{0.077} & 7.77 & \textbf{1.86} & \textbf{0.503} & \textbf{0.025} & \textbf{0.914}\\
            \bottomrule
        \end{tabular}
    }
    \caption{Quantitative comparison for self-reenactment averaged over 2 persons from the Multiface dataset.}
    \label{tab:rebuttal_results_multiface}
\end{table}

%% file: main.bbl
\begin{thebibliography}{95}
\providecommand{\natexlab}[1]{#1}
\providecommand{\url}[1]{\texttt{#1}}
\expandafter\ifx\csname urlstyle\endcsname\relax
  \providecommand{\doi}[1]{doi: #1}\else
  \providecommand{\doi}{doi: \begingroup \urlstyle{rm}\Url}\fi

\bibitem[Athar et~al.(2022)Athar, Xu, Sunkavalli, Shechtman, and Shu]{athar2022rignerf}
ShahRukh Athar, Zexiang Xu, Kalyan Sunkavalli, Eli Shechtman, and Zhixin Shu.
\newblock Rignerf: Fully controllable neural 3d portraits.
\newblock In \emph{Proceedings of the IEEE/CVF conference on Computer Vision and Pattern Recognition}, pages 20364--20373, 2022.

\bibitem[Bergman et~al.(2022)Bergman, Kellnhofer, Yifan, Chan, Lindell, and Wetzstein]{bergman2022gnarf}
Alexander Bergman, Petr Kellnhofer, Wang Yifan, Eric Chan, David Lindell, and Gordon Wetzstein.
\newblock Generative neural articulated radiance fields.
\newblock \emph{Advances in Neural Information Processing Systems}, 35:\penalty0 19900--19916, 2022.

\bibitem[Bergman et~al.(2023)Bergman, Yifan, and Wetzstein]{bergman2023articulated3DHead}
Alexander~W Bergman, Wang Yifan, and Gordon Wetzstein.
\newblock Articulated 3d head avatar generation using text-to-image diffusion models.
\newblock \emph{arXiv preprint arXiv:2307.04859}, 2023.

\bibitem[Blanz and Vetter(2023)]{blanz2023morphable}
Volker Blanz and Thomas Vetter.
\newblock A morphable model for the synthesis of 3d faces.
\newblock In \emph{Seminal Graphics Papers: Pushing the Boundaries, Volume 2}, pages 157--164. 2023.

\bibitem[Blattmann et~al.(2023)Blattmann, Rombach, Ling, Dockhorn, Kim, Fidler, and Kreis]{blattmann2023align}
Andreas Blattmann, Robin Rombach, Huan Ling, Tim Dockhorn, Seung~Wook Kim, Sanja Fidler, and Karsten Kreis.
\newblock Align your latents: High-resolution video synthesis with latent diffusion models.
\newblock In \emph{Proceedings of the IEEE/CVF Conference on Computer Vision and Pattern Recognition}, pages 22563--22575, 2023.

\bibitem[Cao et~al.(2023)Cao, Cao, Han, Shan, and Wong]{cao2023dreamavatar}
Yukang Cao, Yan-Pei Cao, Kai Han, Ying Shan, and Kwan-Yee~K Wong.
\newblock Dreamavatar: Text-and-shape guided 3d human avatar generation via diffusion models.
\newblock \emph{arXiv preprint arXiv:2304.00916}, 2023.

\bibitem[Chai et~al.(2023)Chai, Zhang, He, Tan, Baltrusaitis, Wu, Li, Zhao, Yuan, and Bian]{Chai2023HiFace}
Zenghao Chai, Tianke Zhang, Tianyu He, Xu Tan, Tadas Baltrusaitis, HsiangTao Wu, Runnan Li, Sheng Zhao, Chun Yuan, and Jiang Bian.
\newblock Hiface: High-fidelity 3d face reconstruction by learning static and dynamic details.
\newblock In \emph{Proceedings of the IEEE/CVF International Conference on Computer Vision (ICCV)}, pages 9087--9098, 2023.

\bibitem[Chan et~al.(2022)Chan, Lin, Chan, Nagano, Pan, De~Mello, Gallo, Guibas, Tremblay, Khamis, et~al.]{chan2022eg3d}
Eric~R Chan, Connor~Z Lin, Matthew~A Chan, Koki Nagano, Boxiao Pan, Shalini De~Mello, Orazio Gallo, Leonidas~J Guibas, Jonathan Tremblay, Sameh Khamis, et~al.
\newblock Efficient geometry-aware 3d generative adversarial networks.
\newblock In \emph{Proceedings of the IEEE/CVF Conference on Computer Vision and Pattern Recognition}, pages 16123--16133, 2022.

\bibitem[Chu et~al.(2024)Chu, Li, Zeng, Yang, Lin, Liu, and Harada]{chu2024gpavatar}
Xuangeng Chu, Yu Li, Ailing Zeng, Tianyu Yang, Lijian Lin, Yunfei Liu, and Tatsuya Harada.
\newblock {GPA}vatar: Generalizable and precise head avatar from image(s).
\newblock In \emph{The Twelfth International Conference on Learning Representations}, 2024.

\bibitem[Deng et~al.(2019)Deng, Guo, Xue, and Zafeiriou]{deng2019arcface}
Jiankang Deng, Jia Guo, Niannan Xue, and Stefanos Zafeiriou.
\newblock Arcface: Additive angular margin loss for deep face recognition.
\newblock In \emph{Proceedings of the IEEE/CVF conference on computer vision and pattern recognition}, pages 4690--4699, 2019.

\bibitem[Deng et~al.(2020)Deng, Yang, Chen, Wen, and Tong]{deng2020discofacegan}
Yu Deng, Jiaolong Yang, Dong Chen, Fang Wen, and Xin Tong.
\newblock Disentangled and controllable face image generation via 3d imitative-contrastive learning.
\newblock In \emph{IEEE Computer Vision and Pattern Recognition}, 2020.

\bibitem[Ding et~al.(2023)Ding, Zhang, Xia, Jebe, Tu, and Zhang]{ding2023diffusionrig}
Zheng Ding, Xuaner Zhang, Zhihao Xia, Lars Jebe, Zhuowen Tu, and Xiuming Zhang.
\newblock Diffusionrig: Learning personalized priors for facial appearance editing.
\newblock In \emph{Proceedings of the IEEE/CVF Conference on Computer Vision and Pattern Recognition}, pages 12736--12746, 2023.

\bibitem[Esser et~al.(2021)Esser, Rombach, and Ommer]{esser2021taming}
Patrick Esser, Robin Rombach, and Bjorn Ommer.
\newblock Taming transformers for high-resolution image synthesis.
\newblock In \emph{Proceedings of the IEEE/CVF conference on computer vision and pattern recognition}, pages 12873--12883, 2021.

\bibitem[Feng et~al.(2021)Feng, Feng, Black, and Bolkart]{DECA:Siggraph2021}
Yao Feng, Haiwen Feng, Michael~J. Black, and Timo Bolkart.
\newblock Learning an animatable detailed {3D} face model from in-the-wild images.
\newblock 2021.

\bibitem[Fi{\v{s}}er et~al.(2017)Fi{\v{s}}er, Jamri{\v{s}}ka, Simons, Shechtman, Lu, Asente, Luk{\'a}{\v{c}}, and S{\`y}kora]{fivser2017examplefaceanimation}
Jakub Fi{\v{s}}er, Ond{\v{r}}ej Jamri{\v{s}}ka, David Simons, Eli Shechtman, Jingwan Lu, Paul Asente, Michal Luk{\'a}{\v{c}}, and Daniel S{\`y}kora.
\newblock Example-based synthesis of stylized facial animations.
\newblock \emph{ACM Transactions on Graphics (TOG)}, 36\penalty0 (4):\penalty0 1--11, 2017.

\bibitem[Gafni et~al.(2021)Gafni, Thies, Zollhofer, and Nie{\ss}ner]{gafni2021nerface}
Guy Gafni, Justus Thies, Michael Zollhofer, and Matthias Nie{\ss}ner.
\newblock Dynamic neural radiance fields for monocular 4d facial avatar reconstruction.
\newblock In \emph{Proceedings of the IEEE/CVF Conference on Computer Vision and Pattern Recognition}, pages 8649--8658, 2021.

\bibitem[Geyer et~al.(2023)Geyer, Bar-Tal, Bagon, and Dekel]{geyer2023tokenflow}
Michal Geyer, Omer Bar-Tal, Shai Bagon, and Tali Dekel.
\newblock Tokenflow: Consistent diffusion features for consistent video editing.
\newblock \emph{arXiv preprint arXiv:2307.10373}, 2023.

\bibitem[Giebenhain et~al.(2023)Giebenhain, Kirschstein, Georgopoulos, R{\"u}nz, Agapito, and Nie{\ss}ner]{giebenhain2023nphm}
Simon Giebenhain, Tobias Kirschstein, Markos Georgopoulos, Martin R{\"u}nz, Lourdes Agapito, and Matthias Nie{\ss}ner.
\newblock Learning neural parametric head models.
\newblock In \emph{Proceedings of the IEEE/CVF Conference on Computer Vision and Pattern Recognition}, pages 21003--21012, 2023.

\bibitem[Giebenhain et~al.(2024)Giebenhain, Kirschstein, Georgopoulos, R{\"{u}}nz, Agapito, and Nie{\ss}ner]{giebenhain2024mononphm}
Simon Giebenhain, Tobias Kirschstein, Markos Georgopoulos, Martin R{\"{u}}nz, Lourdes Agapito, and Matthias Nie{\ss}ner.
\newblock Mononphm: Dynamic head reconstruction from monocular videos.
\newblock In \emph{Proc. IEEE Conf. on Computer Vision and Pattern Recognition (CVPR)}, 2024.

\bibitem[Goodfellow et~al.(2020)Goodfellow, Pouget-Abadie, Mirza, Xu, Warde-Farley, Ozair, Courville, and Bengio]{goodfellow2020gan}
Ian Goodfellow, Jean Pouget-Abadie, Mehdi Mirza, Bing Xu, David Warde-Farley, Sherjil Ozair, Aaron Courville, and Yoshua Bengio.
\newblock Generative adversarial networks.
\newblock \emph{Communications of the ACM}, 63\penalty0 (11):\penalty0 139--144, 2020.

\bibitem[Grassal et~al.(2022)Grassal, Prinzler, Leistner, Rother, Nie{\ss}ner, and Thies]{grassal2022nha}
Philip-William Grassal, Malte Prinzler, Titus Leistner, Carsten Rother, Matthias Nie{\ss}ner, and Justus Thies.
\newblock Neural head avatars from monocular rgb videos.
\newblock In \emph{Proceedings of the IEEE/CVF Conference on Computer Vision and Pattern Recognition}, pages 18653--18664, 2022.

\bibitem[Gu et~al.(2023)Gu, Gao, Zhai, Chen, Liu, and Susskind]{gu2023controllable3DDiffusion}
Jiatao Gu, Qingzhe Gao, Shuangfei Zhai, Baoquan Chen, Lingjie Liu, and Josh Susskind.
\newblock Learning controllable 3d diffusion models from single-view images.
\newblock \emph{arXiv preprint arXiv:2304.06700}, 2023.

\bibitem[Ho et~al.(2020)Ho, Jain, and Abbeel]{ho2020ddpm}
Jonathan Ho, Ajay Jain, and Pieter Abbeel.
\newblock Denoising diffusion probabilistic models.
\newblock \emph{Advances in neural information processing systems}, 33:\penalty0 6840--6851, 2020.

\bibitem[Ho et~al.(2022{\natexlab{a}})Ho, Chan, Saharia, Whang, Gao, Gritsenko, Kingma, Poole, Norouzi, Fleet, et~al.]{ho2022imagenvideo}
Jonathan Ho, William Chan, Chitwan Saharia, Jay Whang, Ruiqi Gao, Alexey Gritsenko, Diederik~P Kingma, Ben Poole, Mohammad Norouzi, David~J Fleet, et~al.
\newblock Imagen video: High definition video generation with diffusion models.
\newblock \emph{arXiv preprint arXiv:2210.02303}, 2022{\natexlab{a}}.

\bibitem[Ho et~al.(2022{\natexlab{b}})Ho, Salimans, Gritsenko, Chan, Norouzi, and Fleet]{ho2022vdm}
Jonathan Ho, Tim Salimans, Alexey Gritsenko, William Chan, Mohammad Norouzi, and David~J Fleet.
\newblock Video diffusion models.
\newblock \emph{arXiv:2204.03458}, 2022{\natexlab{b}}.

\bibitem[Hong et~al.(2022)Hong, Zhang, Shen, and Xu]{hong2022dagan}
Fa-Ting Hong, Longhao Zhang, Li Shen, and Dan Xu.
\newblock Depth-aware generative adversarial network for talking head video generation.
\newblock In \emph{Proceedings of the IEEE/CVF conference on computer vision and pattern recognition}, pages 3397--3406, 2022.

\bibitem[Huang et~al.(2023)Huang, Chan, Jiang, and Liu]{huang2023collaborativediffusionfaces}
Ziqi Huang, Kelvin~C.K. Chan, Yuming Jiang, and Ziwei Liu.
\newblock Collaborative diffusion for multi-modal face generation and editing.
\newblock In \emph{Proceedings of the IEEE/CVF Conference on Computer Vision and Pattern Recognition}, 2023.

\bibitem[Jamri{\v{s}}ka et~al.(2019)Jamri{\v{s}}ka, Sochorov{\'a}, Texler, Luk{\'a}{\v{c}}, Fi{\v{s}}er, Lu, Shechtman, and S{\`y}kora]{jamrivska2019stylizingvideoexample}
Ond{\v{r}}ej Jamri{\v{s}}ka, {\v{S}}{\'a}rka Sochorov{\'a}, Ond{\v{r}}ej Texler, Michal Luk{\'a}{\v{c}}, Jakub Fi{\v{s}}er, Jingwan Lu, Eli Shechtman, and Daniel S{\`y}kora.
\newblock Stylizing video by example.
\newblock \emph{ACM Transactions on Graphics (TOG)}, 38\penalty0 (4):\penalty0 1--11, 2019.

\bibitem[Jiang et~al.(2023)Jiang, Wang, Zhang, Chai, He, Chen, and Liao]{jiang2023avatarcraft}
Ruixiang Jiang, Can Wang, Jingbo Zhang, Menglei Chai, Mingming He, Dongdong Chen, and Jing Liao.
\newblock Avatarcraft: Transforming text into neural human avatars with parameterized shape and pose control.
\newblock \emph{arXiv preprint arXiv:2303.17606}, 2023.

\bibitem[Jin et~al.(2021)Jin, Liao, and Shao]{jin2021pipnet}
Haibo Jin, Shengcai Liao, and Ling Shao.
\newblock Pixel-in-pixel net: Towards efficient facial landmark detection in the wild.
\newblock \emph{International Journal of Computer Vision}, 2021.

\bibitem[Karras et~al.(2019)Karras, Laine, and Aila]{karras2019stylegan}
Tero Karras, Samuli Laine, and Timo Aila.
\newblock A style-based generator architecture for generative adversarial networks.
\newblock In \emph{Proceedings of the IEEE/CVF conference on computer vision and pattern recognition}, pages 4401--4410, 2019.

\bibitem[Karras et~al.(2021)Karras, Aittala, Laine, H{\"a}rk{\"o}nen, Hellsten, Lehtinen, and Aila]{karras2021stylegan3}
Tero Karras, Miika Aittala, Samuli Laine, Erik H{\"a}rk{\"o}nen, Janne Hellsten, Jaakko Lehtinen, and Timo Aila.
\newblock Alias-free generative adversarial networks.
\newblock \emph{Advances in neural information processing systems}, 34:\penalty0 852--863, 2021.

\bibitem[Khakhulin et~al.(2022)Khakhulin, Sklyarova, Lempitsky, and Zakharov]{khakhulin2022rome}
Taras Khakhulin, Vanessa Sklyarova, Victor Lempitsky, and Egor Zakharov.
\newblock Realistic one-shot mesh-based head avatars.
\newblock In \emph{European Conference on Computer Vision}, pages 345--362. Springer, 2022.

\bibitem[Kim et~al.(2018)Kim, Garrido, Tewari, Xu, Thies, Nie{\ss}ner, P{\'e}rez, Richardt, Zoll{\"o}fer, and Theobalt]{kim2018dvp}
Hyeongwoo Kim, Pablo Garrido, Ayush Tewari, Weipeng Xu, Justus Thies, Matthias Nie{\ss}ner, Patrick P{\'e}rez, Christian Richardt, Michael Zoll{\"o}fer, and Christian Theobalt.
\newblock Deep video portraits.
\newblock \emph{ACM Transactions on Graphics (TOG)}, 37\penalty0 (4):\penalty0 163, 2018.

\bibitem[Kingma and Ba(2014)]{kingma2014adam}
Diederik~P Kingma and Jimmy Ba.
\newblock Adam: A method for stochastic optimization.
\newblock \emph{arXiv preprint arXiv:1412.6980}, 2014.

\bibitem[Kirschstein et~al.(2023)Kirschstein, Qian, Giebenhain, Walter, and Nie\ss{}ner]{kirschstein2023nersemble}
Tobias Kirschstein, Shenhan Qian, Simon Giebenhain, Tim Walter, and Matthias Nie\ss{}ner.
\newblock Nersemble: Multi-view radiance field reconstruction of human heads.
\newblock \emph{ACM Trans. Graph.}, 42\penalty0 (4), 2023.

\bibitem[Laine et~al.(2020)Laine, Hellsten, Karras, Seol, Lehtinen, and Aila]{Laine2020nvdiffrast}
Samuli Laine, Janne Hellsten, Tero Karras, Yeongho Seol, Jaakko Lehtinen, and Timo Aila.
\newblock Modular primitives for high-performance differentiable rendering.
\newblock \emph{ACM Transactions on Graphics}, 39\penalty0 (6), 2020.

\bibitem[Li et~al.(2017)Li, Bolkart, Black, Li, and Romero]{FLAME:SiggraphAsia2017}
Tianye Li, Timo Bolkart, Michael.~J. Black, Hao Li, and Javier Romero.
\newblock Learning a model of facial shape and expression from {4D} scans.
\newblock \emph{ACM Transactions on Graphics, (Proc. SIGGRAPH Asia)}, 36\penalty0 (6):\penalty0 194:1--194:17, 2017.

\bibitem[Lin et~al.(2022)Lin, Lindell, Chan, and Wetzstein]{lin2022face3dganinversion}
C.Z. Lin, D.B. Lindell, E.R. Chan, and G. Wetzstein.
\newblock 3d gan inversion for controllable portrait image animation.
\newblock In \emph{ECCV Workshop on Learning to Generate 3D Shapes and Scenes}, 2022.

\bibitem[Lin et~al.(2023{\natexlab{a}})Lin, Nagano, Kautz, Chan, Iqbal, Guibas, Wetzstein, and Khamis]{lin2023ssif}
Connor~Z. Lin, Koki Nagano, Jan Kautz, Eric~R. Chan, Umar Iqbal, Leonidas Guibas, Gordon Wetzstein, and Sameh Khamis.
\newblock Single-shot implicit morphable faces with consistent texture parameterization.
\newblock In \emph{ACM SIGGRAPH 2023 Conference Proceedings}, 2023{\natexlab{a}}.

\bibitem[Lin et~al.(2021)Lin, Ryabtsev, Sengupta, Curless, Seitz, and Kemelmacher-Shlizerman]{lin2021backgroundmatting}
Shanchuan Lin, Andrey Ryabtsev, Soumyadip Sengupta, Brian~L Curless, Steven~M Seitz, and Ira Kemelmacher-Shlizerman.
\newblock Real-time high-resolution background matting.
\newblock In \emph{Proceedings of the IEEE/CVF Conference on Computer Vision and Pattern Recognition}, pages 8762--8771, 2021.

\bibitem[Lin et~al.(2023{\natexlab{b}})Lin, Liu, Li, and Yang]{lin2023diffusionflawed}
Shanchuan Lin, Bingchen Liu, Jiashi Li, and Xiao Yang.
\newblock Common diffusion noise schedules and sample steps are flawed.
\newblock \emph{arXiv preprint arXiv:2305.08891}, 2023{\natexlab{b}}.

\bibitem[Liu et~al.(2023)Liu, Wu, Hoorick, Tokmakov, Zakharov, and Vondrick]{liu2023zero1to3}
Ruoshi Liu, Rundi Wu, Basile~Van Hoorick, Pavel Tokmakov, Sergey Zakharov, and Carl Vondrick.
\newblock Zero-1-to-3: Zero-shot one image to 3d object, 2023.

\bibitem[Lombardi et~al.(2021)Lombardi, Simon, Schwartz, Zollhoefer, Sheikh, and Saragih]{lombardi2021mixture}
Stephen Lombardi, Tomas Simon, Gabriel Schwartz, Michael Zollhoefer, Yaser Sheikh, and Jason Saragih.
\newblock Mixture of volumetric primitives for efficient neural rendering.
\newblock \emph{ACM Transactions on Graphics (ToG)}, 40\penalty0 (4):\penalty0 1--13, 2021.

\bibitem[Loper et~al.(2023)Loper, Mahmood, Romero, Pons-Moll, and Black]{loper2023smpl}
Matthew Loper, Naureen Mahmood, Javier Romero, Gerard Pons-Moll, and Michael~J Black.
\newblock Smpl: A skinned multi-person linear model.
\newblock In \emph{Seminal Graphics Papers: Pushing the Boundaries, Volume 2}, pages 851--866. 2023.

\bibitem[Lorensen and Cline(1998)]{lorensen1998marchingcubes}
William~E Lorensen and Harvey~E Cline.
\newblock Marching cubes: A high resolution 3d surface construction algorithm.
\newblock In \emph{Seminal graphics: pioneering efforts that shaped the field}, pages 347--353. 1998.

\bibitem[Mantiuk et~al.(2021)Mantiuk, Denes, Chapiro, Kaplanyan, Rufo, Bachy, Lian, and Patney]{jod_metric}
Rafa\l{}~K. Mantiuk, Gyorgy Denes, Alexandre Chapiro, Anton Kaplanyan, Gizem Rufo, Romain Bachy, Trisha Lian, and Anjul Patney.
\newblock Fovvideovdp: A visible difference predictor for wide field-of-view video.
\newblock \emph{ACM Trans. Graph.}, 40\penalty0 (4), 2021.

\bibitem[Meng et~al.(2023)Meng, Rombach, Gao, Kingma, Ermon, Ho, and Salimans]{meng2023distillation}
Chenlin Meng, Robin Rombach, Ruiqi Gao, Diederik Kingma, Stefano Ermon, Jonathan Ho, and Tim Salimans.
\newblock On distillation of guided diffusion models.
\newblock In \emph{Proceedings of the IEEE/CVF Conference on Computer Vision and Pattern Recognition}, pages 14297--14306, 2023.

\bibitem[Mensah et~al.(2023)Mensah, Kim, Aittala, Laine, and Lehtinen]{mensah2023hybridgeneratorarchitecture}
Dann Mensah, Nam~Hee Kim, Miika Aittala, Samuli Laine, and Jaakko Lehtinen.
\newblock A hybrid generator architecture for controllable face synthesis.
\newblock In \emph{ACM SIGGRAPH 2023 Conference Proceedings}, pages 1--10, 2023.

\bibitem[Mou et~al.(2023)Mou, Wang, Xie, Wu, Zhang, Qi, Shan, and Qie]{mou2023t2i}
Chong Mou, Xintao Wang, Liangbin Xie, Yanze Wu, Jian Zhang, Zhongang Qi, Ying Shan, and Xiaohu Qie.
\newblock T2i-adapter: Learning adapters to dig out more controllable ability for text-to-image diffusion models.
\newblock \emph{arXiv preprint arXiv:2302.08453}, 2023.

\bibitem[M{\"u}ller et~al.(2023)M{\"u}ller, Siddiqui, Porzi, Bulo, Kontschieder, and Nie{\ss}ner]{muller2023diffrf}
Norman M{\"u}ller, Yawar Siddiqui, Lorenzo Porzi, Samuel~Rota Bulo, Peter Kontschieder, and Matthias Nie{\ss}ner.
\newblock Diffrf: Rendering-guided 3d radiance field diffusion.
\newblock In \emph{Proceedings of the IEEE/CVF Conference on Computer Vision and Pattern Recognition}, pages 4328--4338, 2023.

\bibitem[Nitzan et~al.(2022)Nitzan, Aberman, He, Liba, Yarom, Gandelsman, Mosseri, Pritch, and Cohen-Or]{nitzan2022mystyle}
Yotam Nitzan, Kfir Aberman, Qiurui He, Orly Liba, Michal Yarom, Yossi Gandelsman, Inbar Mosseri, Yael Pritch, and Daniel Cohen-Or.
\newblock Mystyle: A personalized generative prior.
\newblock \emph{arXiv preprint arXiv:2203.17272}, 2022.

\bibitem[Pan et~al.(2023)Pan, Elgharib, Teotia, Tewari, Golyanik, Kortylewski, Theobalt, et~al.]{pan2023avatarstudio}
Mohit~Mendiratta Pan, Mohamed Elgharib, Kartik Teotia, Ayush Tewari, Vladislav Golyanik, Adam Kortylewski, Christian Theobalt, et~al.
\newblock Avatarstudio: Text-driven editing of 3d dynamic human head avatars.
\newblock \emph{arXiv preprint arXiv:2306.00547}, 2023.

\bibitem[Park et~al.(2021)Park, Sinha, Hedman, Barron, Bouaziz, Goldman, Martin-Brualla, and Seitz]{park2021hypernerf}
Keunhong Park, Utkarsh Sinha, Peter Hedman, Jonathan~T Barron, Sofien Bouaziz, Dan~B Goldman, Ricardo Martin-Brualla, and Steven~M Seitz.
\newblock Hypernerf: A higher-dimensional representation for topologically varying neural radiance fields.
\newblock \emph{arXiv preprint arXiv:2106.13228}, 2021.

\bibitem[Paysan et~al.(2009)Paysan, Knothe, Amberg, Romdhani, and Vetter]{paysan2009bfm}
Pascal Paysan, Reinhard Knothe, Brian Amberg, Sami Romdhani, and Thomas Vetter.
\newblock A 3d face model for pose and illumination invariant face recognition.
\newblock In \emph{2009 sixth IEEE international conference on advanced video and signal based surveillance}, pages 296--301. Ieee, 2009.

\bibitem[Poole et~al.(2022)Poole, Jain, Barron, and Mildenhall]{poole2022dreamfusion}
Ben Poole, Ajay Jain, Jonathan~T. Barron, and Ben Mildenhall.
\newblock Dreamfusion: Text-to-3d using 2d diffusion.
\newblock \emph{arXiv}, 2022.

\bibitem[Qin et~al.(2023)Qin, Ye, Yu, Tang, and Zhuang]{qin2023dancingavatar}
Bosheng Qin, Wentao Ye, Qifan Yu, Siliang Tang, and Yueting Zhuang.
\newblock Dancing avatar: Pose and text-guided human motion videos synthesis with image diffusion model.
\newblock \emph{arXiv preprint arXiv:2308.07749}, 2023.

\bibitem[Richardson et~al.(2021)Richardson, Alaluf, Patashnik, Nitzan, Azar, Shapiro, and Cohen-Or]{richardson2021pixelstylepixel}
Elad Richardson, Yuval Alaluf, Or Patashnik, Yotam Nitzan, Yaniv Azar, Stav Shapiro, and Daniel Cohen-Or.
\newblock Encoding in style: a stylegan encoder for image-to-image translation.
\newblock In \emph{IEEE/CVF Conference on Computer Vision and Pattern Recognition (CVPR)}, 2021.

\bibitem[Rombach et~al.(2022)Rombach, Blattmann, Lorenz, Esser, and Ommer]{rombach2022ldm}
Robin Rombach, Andreas Blattmann, Dominik Lorenz, Patrick Esser, and Bj{\"o}rn Ommer.
\newblock High-resolution image synthesis with latent diffusion models.
\newblock In \emph{Proceedings of the IEEE/CVF conference on computer vision and pattern recognition}, pages 10684--10695, 2022.

\bibitem[R{\"o}ssler et~al.(2018)R{\"o}ssler, Cozzolino, Verdoliva, Riess, Thies, and Nie{\ss}ner]{rossler2018faceforensics}
Andreas R{\"o}ssler, Davide Cozzolino, Luisa Verdoliva, Christian Riess, Justus Thies, and Matthias Nie{\ss}ner.
\newblock Faceforensics: A large-scale video dataset for forgery detection in human faces.
\newblock \emph{arXiv preprint arXiv:1803.09179}, 2018.

\bibitem[Salimans and Ho(2022)]{salimans2022progressivedistillation}
Tim Salimans and Jonathan Ho.
\newblock Progressive distillation for fast sampling of diffusion models.
\newblock \emph{arXiv preprint arXiv:2202.00512}, 2022.

\bibitem[Sch\"{o}nberger and Frahm(2016)]{schoenberger2016sfm_colmap}
Johannes~Lutz Sch\"{o}nberger and Jan-Michael Frahm.
\newblock Structure-from-motion revisited.
\newblock In \emph{Conference on Computer Vision and Pattern Recognition (CVPR)}, 2016.

\bibitem[Sch\"{o}nberger et~al.(2016)Sch\"{o}nberger, Zheng, Pollefeys, and Frahm]{schoenberger2016mvs_colmap}
Johannes~Lutz Sch\"{o}nberger, Enliang Zheng, Marc Pollefeys, and Jan-Michael Frahm.
\newblock Pixelwise view selection for unstructured multi-view stereo.
\newblock In \emph{European Conference on Computer Vision (ECCV)}, 2016.

\bibitem[Singer et~al.(2022)Singer, Polyak, Hayes, Yin, An, Zhang, Hu, Yang, Ashual, Gafni, et~al.]{singer2022makeavideo}
Uriel Singer, Adam Polyak, Thomas Hayes, Xi Yin, Jie An, Songyang Zhang, Qiyuan Hu, Harry Yang, Oron Ashual, Oran Gafni, et~al.
\newblock Make-a-video: Text-to-video generation without text-video data.
\newblock \emph{arXiv preprint arXiv:2209.14792}, 2022.

\bibitem[Song et~al.(2023)Song, Dhariwal, Chen, and Sutskever]{song2023consistency}
Yang Song, Prafulla Dhariwal, Mark Chen, and Ilya Sutskever.
\newblock Consistency models.
\newblock 2023.

\bibitem[Stypu{\l}kowski et~al.(2023)Stypu{\l}kowski, Vougioukas, He, Zieba, Petridis, and Pantic]{stypulkowski2022diffusedheads}
Micha{\l} Stypu{\l}kowski, Konstantinos Vougioukas, Sen He, Maciej Zieba, Stavros Petridis, and Maja Pantic.
\newblock Diffused heads: Diffusion models beat gans on talking-face generation.
\newblock \emph{arXiv preprint arXiv:2301.03396}, 2023.

\bibitem[Sun et~al.(2023)Sun, Wang, Wang, Li, Zhang, Zhang, and Liu]{sun2023next3d}
Jingxiang Sun, Xuan Wang, Lizhen Wang, Xiaoyu Li, Yong Zhang, Hongwen Zhang, and Yebin Liu.
\newblock Next3d: Generative neural texture rasterization for 3d-aware head avatars.
\newblock In \emph{Proceedings of the IEEE/CVF Conference on Computer Vision and Pattern Recognition}, pages 20991--21002, 2023.

\bibitem[Svitov et~al.(2023)Svitov, Gudkov, Bashirov, and Lempitsky]{svitov2023dinar}
David Svitov, Dmitrii Gudkov, Renat Bashirov, and Victor Lempitsky.
\newblock Dinar: Diffusion inpainting of neural textures for one-shot human avatars.
\newblock In \emph{Proceedings of the IEEE/CVF International Conference on Computer Vision}, pages 7062--7072, 2023.

\bibitem[Tang et~al.(2023)Tang, Zhang, Yang, Zhang, Chen, Ma, and Wen]{tang20233dfaceshop}
Junshu Tang, Bo Zhang, Binxin Yang, Ting Zhang, Dong Chen, Lizhuang Ma, and Fang Wen.
\newblock 3dfaceshop: Explicitly controllable 3d-aware portrait generation.
\newblock \emph{IEEE Transactions on Visualization and Computer Graphics}, 2023.

\bibitem[Tewari et~al.(2020{\natexlab{a}})Tewari, Elgharib, {B R}, Bernard, Seidel, P{\'e}rez, Z{\"o}llhofer, and Theobalt]{tewari2020pie}
Ayush Tewari, Mohamed Elgharib, Mallikarjun {B R}, Florian Bernard, Hans-Peter Seidel, Patrick P{\'e}rez, Michael Z{\"o}llhofer, and Christian Theobalt.
\newblock Pie: Portrait image embedding for semantic control.
\newblock 2020{\natexlab{a}}.

\bibitem[Tewari et~al.(2020{\natexlab{b}})Tewari, Elgharib, Bharaj, Bernard, Seidel, P{\'e}rez, Z{\"o}llhofer, and Theobalt]{tewari2020stylerig}
Ayush Tewari, Mohamed Elgharib, Gaurav Bharaj, Florian Bernard, Hans-Peter Seidel, Patrick P{\'e}rez, Michael Z{\"o}llhofer, and Christian Theobalt.
\newblock Stylerig: Rigging stylegan for 3d control over portrait images, cvpr 2020.
\newblock In \emph{{IEEE} Conference on Computer Vision and Pattern Recognition (CVPR)}. {IEEE}, 2020{\natexlab{b}}.

\bibitem[Thies et~al.(2019)Thies, Zollh{\"o}fer, and Nie{\ss}ner]{thies2019dnr}
Justus Thies, Michael Zollh{\"o}fer, and Matthias Nie{\ss}ner.
\newblock Deferred neural rendering: Image synthesis using neural textures.
\newblock \emph{ACM Transactions on Graphics 2019 (TOG)}, 2019.

\bibitem[Tov et~al.(2021)Tov, Alaluf, Nitzan, Patashnik, and Cohen-Or]{tov2021styleganmanipulation}
Omer Tov, Yuval Alaluf, Yotam Nitzan, Or Patashnik, and Daniel Cohen-Or.
\newblock Designing an encoder for stylegan image manipulation.
\newblock \emph{ACM Transactions on Graphics (TOG)}, 40\penalty0 (4):\penalty0 1--14, 2021.

\bibitem[Wang et~al.(2022)Wang, Chen, Yu, Ma, Li, and Liu]{wang2022faceverse}
Lizhen Wang, Zhiyua Chen, Tao Yu, Chenguang Ma, Liang Li, and Yebin Liu.
\newblock Faceverse: a fine-grained and detail-controllable 3d face morphable model from a hybrid dataset.
\newblock In \emph{IEEE Conference on Computer Vision and Pattern Recognition (CVPR2022)}, 2022.

\bibitem[Wang et~al.(2023{\natexlab{a}})Wang, Zhang, Zhang, Gu, Bao, Baltrusaitis, Shen, Chen, Wen, Chen, et~al.]{wang2023rodin}
Tengfei Wang, Bo Zhang, Ting Zhang, Shuyang Gu, Jianmin Bao, Tadas Baltrusaitis, Jingjing Shen, Dong Chen, Fang Wen, Qifeng Chen, et~al.
\newblock Rodin: A generative model for sculpting 3d digital avatars using diffusion.
\newblock In \emph{Proceedings of the IEEE/CVF Conference on Computer Vision and Pattern Recognition}, pages 4563--4573, 2023{\natexlab{a}}.

\bibitem[Wang et~al.(2021)Wang, Mallya, and Liu]{wang2021facevid2vid}
Ting-Chun Wang, Arun Mallya, and Ming-Yu Liu.
\newblock One-shot free-view neural talking-head synthesis for video conferencing.
\newblock In \emph{Proceedings of the IEEE Conference on Computer Vision and Pattern Recognition}, 2021.

\bibitem[Wang et~al.(2023{\natexlab{b}})Wang, Chen, Ma, Zhou, Huang, Wang, Yang, He, Yu, Yang, et~al.]{wang2023lavie}
Yaohui Wang, Xinyuan Chen, Xin Ma, Shangchen Zhou, Ziqi Huang, Yi Wang, Ceyuan Yang, Yinan He, Jiashuo Yu, Peiqing Yang, et~al.
\newblock Lavie: High-quality video generation with cascaded latent diffusion models.
\newblock \emph{arXiv preprint arXiv:2309.15103}, 2023{\natexlab{b}}.

\bibitem[Wu et~al.(2023)Wu, Ge, Wang, Lei, Gu, Shi, Hsu, Shan, Qie, and Shou]{wu2023tuneavideo}
Jay~Zhangjie Wu, Yixiao Ge, Xintao Wang, Stan~Weixian Lei, Yuchao Gu, Yufei Shi, Wynne Hsu, Ying Shan, Xiaohu Qie, and Mike~Zheng Shou.
\newblock Tune-a-video: One-shot tuning of image diffusion models for text-to-video generation.
\newblock In \emph{Proceedings of the IEEE/CVF International Conference on Computer Vision}, pages 7623--7633, 2023.

\bibitem[Wu et~al.(2022)Wu, Deng, Yang, Wei, Qifeng, and Tong]{yue2022anifacegan}
Yue Wu, Yu Deng, Jiaolong Yang, Fangyun Wei, Chen Qifeng, and Xin Tong.
\newblock Anifacegan: Animatable 3d-aware face image generation for video avatars.
\newblock In \emph{Advances in Neural Information Processing Systems}, 2022.

\bibitem[Wuu et~al.(2022)Wuu, Zheng, Ardisson, Bali, Belko, Brockmeyer, Evans, Godisart, Ha, Huang, Hypes, Koska, Krenn, Lombardi, Luo, McPhail, Millerschoen, Perdoch, Pitts, Richard, Saragih, Saragih, Shiratori, Simon, Stewart, Trimble, Weng, Whitewolf, Wu, Yu, and Sheikh]{wuu2022multiface}
Cheng-hsin Wuu, Ningyuan Zheng, Scott Ardisson, Rohan Bali, Danielle Belko, Eric Brockmeyer, Lucas Evans, Timothy Godisart, Hyowon Ha, Xuhua Huang, Alexander Hypes, Taylor Koska, Steven Krenn, Stephen Lombardi, Xiaomin Luo, Kevyn McPhail, Laura Millerschoen, Michal Perdoch, Mark Pitts, Alexander Richard, Jason Saragih, Junko Saragih, Takaaki Shiratori, Tomas Simon, Matt Stewart, Autumn Trimble, Xinshuo Weng, David Whitewolf, Chenglei Wu, Shoou-I Yu, and Yaser Sheikh.
\newblock Multiface: A dataset for neural face rendering.
\newblock In \emph{arXiv}, 2022.

\bibitem[Yan et~al.(2023)Yan, Liew, Mai, Lin, and Feng]{yan2023magicprop}
Hanshu Yan, Jun~Hao Liew, Long Mai, Shanchuan Lin, and Jiashi Feng.
\newblock Magicprop: Diffusion-based video editing via motion-aware appearance propagation.
\newblock \emph{arXiv preprint arXiv:2309.00908}, 2023.

\bibitem[Yang et~al.(2020)Yang, Zhu, Wang, Huang, Shen, Yang, and Cao]{yang2020facescape}
Haotian Yang, Hao Zhu, Yanru Wang, Mingkai Huang, Qiu Shen, Ruigang Yang, and Xun Cao.
\newblock Facescape: a large-scale high quality 3d face dataset and detailed riggable 3d face prediction.
\newblock In \emph{Proceedings of the IEEE Conference on Computer Vision and Pattern Recognition (CVPR)}, 2020.

\bibitem[Yang et~al.(2022)Yang, Jiang, Liu, and Loy]{yang2022Vtoonify}
Shuai Yang, Liming Jiang, Ziwei Liu, and Chen~Change Loy.
\newblock Vtoonify: Controllable high-resolution portrait video style transfer.
\newblock \emph{ACM Transactions on Graphics (TOG)}, 41\penalty0 (6):\penalty0 1--15, 2022.

\bibitem[Yang et~al.(2023)Yang, Zhou, Liu, , and Loy]{yang2023rerender}
Shuai Yang, Yifan Zhou, Ziwei Liu, , and Chen~Change Loy.
\newblock Rerender a video: Zero-shot text-guided video-to-video translation.
\newblock In \emph{ACM SIGGRAPH Asia Conference Proceedings}, 2023.

\bibitem[Ye et~al.(2023)Ye, Zhang, Liu, Han, and Yang]{ye2023ipadapter}
Hu Ye, Jun Zhang, Sibo Liu, Xiao Han, and Wei Yang.
\newblock Ip-adapter: Text compatible image prompt adapter for text-to-image diffusion models.
\newblock \emph{arXiv preprint arXiv:2308.06721}, 2023.

\bibitem[Zeng et~al.(2023{\natexlab{a}})Zeng, Liu, Gao, Liu, Li, Liu, and Zhang]{zeng2023attributeguideddiffusion}
Bohan Zeng, Xuhui Liu, Sicheng Gao, Boyu Liu, Hong Li, Jianzhuang Liu, and Baochang Zhang.
\newblock Face animation with an attribute-guided diffusion model.
\newblock In \emph{Proceedings of the IEEE/CVF Conference on Computer Vision and Pattern Recognition}, pages 628--637, 2023{\natexlab{a}}.

\bibitem[Zeng et~al.(2023{\natexlab{b}})Zeng, Lu, Ji, Yao, Zhu, and Cao]{zeng2023avatarbooth}
Yifei Zeng, Yuanxun Lu, Xinya Ji, Yao Yao, Hao Zhu, and Xun Cao.
\newblock Avatarbooth: High-quality and customizable 3d human avatar generation.
\newblock \emph{arXiv preprint arXiv:2306.09864}, 2023{\natexlab{b}}.

\bibitem[Zhang et~al.(2023{\natexlab{a}})Zhang, Chen, Fu, Zhou, Yu, Wang, Fu, Chen, Lin, and Shen]{zhang2023styleavatar3d}
Chi Zhang, Yiwen Chen, Yijun Fu, Zhenglin Zhou, Gang Yu, Billzb Wang, Bin Fu, Tao Chen, Guosheng Lin, and Chunhua Shen.
\newblock Styleavatar3d: Leveraging image-text diffusion models for high-fidelity 3d avatar generation.
\newblock \emph{arXiv preprint arXiv:2305.19012}, 2023{\natexlab{a}}.

\bibitem[Zhang et~al.(2023{\natexlab{b}})Zhang, Rao, and Agrawala]{zhang2023controlnet}
Lvmin Zhang, Anyi Rao, and Maneesh Agrawala.
\newblock Adding conditional control to text-to-image diffusion models, 2023{\natexlab{b}}.

\bibitem[Zhang et~al.(2018)Zhang, Isola, Efros, Shechtman, and Wang]{zhang2018lpips}
Richard Zhang, Phillip Isola, Alexei~A Efros, Eli Shechtman, and Oliver Wang.
\newblock The unreasonable effectiveness of deep features as a perceptual metric.
\newblock In \emph{Proceedings of the IEEE conference on computer vision and pattern recognition}, pages 586--595, 2018.

\bibitem[Zheng et~al.(2021)Zheng, Yang, Zhang, Bao, Chen, Huang, Yuan, Chen, Zeng, and Wen]{zheng2021farl}
Yinglin Zheng, Hao Yang, Ting Zhang, Jianmin Bao, Dongdong Chen, Yangyu Huang, Lu Yuan, Dong Chen, Ming Zeng, and Fang Wen.
\newblock General facial representation learning in a visual-linguistic manner.
\newblock \emph{arXiv preprint arXiv:2112.03109}, 2021.

\bibitem[Zheng et~al.(2022)Zheng, Abrevaya, B{\"u}hler, Chen, Black, and Hilliges]{zheng2022imavatar}
Yufeng Zheng, Victoria~Fern{\'a}ndez Abrevaya, Marcel~C B{\"u}hler, Xu Chen, Michael~J Black, and Otmar Hilliges.
\newblock Im avatar: Implicit morphable head avatars from videos.
\newblock In \emph{Proceedings of the IEEE/CVF Conference on Computer Vision and Pattern Recognition}, pages 13545--13555, 2022.

\bibitem[Zheng et~al.(2023)Zheng, Yifan, Wetzstein, Black, and Hilliges]{zheng2023pointavatar}
Yufeng Zheng, Wang Yifan, Gordon Wetzstein, Michael~J Black, and Otmar Hilliges.
\newblock Pointavatar: Deformable point-based head avatars from videos.
\newblock In \emph{Proceedings of the IEEE/CVF Conference on Computer Vision and Pattern Recognition}, pages 21057--21067, 2023.

\bibitem[Zhu et~al.(2017)Zhu, Park, Isola, and Efros]{zhu2017pix2pix}
Jun-Yan Zhu, Taesung Park, Phillip Isola, and Alexei~A Efros.
\newblock Unpaired image-to-image translation using cycle-consistent adversarial networks.
\newblock In \emph{Proceedings of the IEEE international conference on computer vision}, pages 2223--2232, 2017.

\bibitem[Zielonka et~al.(2023)Zielonka, Bolkart, and Thies]{zielonka2023insta}
Wojciech Zielonka, Timo Bolkart, and Justus Thies.
\newblock Instant volumetric head avatars.
\newblock In \emph{Proceedings of the IEEE/CVF Conference on Computer Vision and Pattern Recognition}, pages 4574--4584, 2023.

\end{thebibliography}
